\documentclass{article}
\usepackage{amssymb}
\usepackage{graphicx}
\usepackage{arydshln}
\usepackage{xcolor}
\usepackage{booktabs}
\usepackage{amsmath}
\usepackage[table,xcdraw]{xcolor}
\usepackage{appendix}

\usepackage{wrapfig}
\usepackage{caption}
\usepackage{tcolorbox}
\tcbuselibrary{listings,skins,breakable}
\usepackage[final]{corl_2025} 

\title{RoboChemist: Long-Horizon and Safety-Compliant Robotic Chemical Experimentation}

\author{
Zongzheng Zhang\footnotemark[1] $^{\ 1}$, Chenghao Yue\footnotemark[1] $^{\ 1}$, Haobo Xu\footnotemark[1] $^{\ 2}$\vspace{0.1cm},\\
\textbf{Minwen Liao$^{1}$, Xianglin Qi$^1$, }
\textbf{Huan-ang Gao$^1$, Ziwei Wang$^3$, Hao Zhao\footnotemark[2] $^{\ 1,2}$}\vspace{0.3cm}\\
\footnotemark[1]\ \ Equal Contribution;  \footnotemark[2]\ \ Corresponding author \vspace{0.3cm}\\
    $^1$ 
  Beijing Academy of Artificial Intelligence, BAAI\\
    $^2$ Institute for AI Industry Research (AIR), Tsinghua Univeristy \\
    $^3$ Nanyang Technological University \vspace{0.15cm}\\
\texttt{zhaohao@air.tsinghua.edu.cn}\vspace{0.15cm}\\
  \href{https://zzongzheng0918.github.io/RoboChemist.github.io/}{https://zzongzheng0918.github.io/RoboChemist.github.io/}
}

\definecolor{forestgreen}{rgb}{0.0, 0.5, 0.0}
\definecolor{darkpastelgreen}{rgb}{0.01, 0.75, 0.24}
\definecolor{darkgreen}{rgb}{0.00, 0.8, 0.2}
\definecolor{darkyellow}{rgb}{0.96, 0.75, 0.00}

\begin{document}
\maketitle

\vspace{-1cm}
\begin{abstract}
Robotic chemists promise to both liberate human experts from repetitive tasks and accelerate scientific discovery, yet remain in their infancy. Chemical experiments involve long-horizon procedures over hazardous and deformable substances, where success requires not only task completion but also strict compliance with experimental norms. To address these challenges, we propose \textit{RoboChemist}, a dual-loop framework that integrates Vision-Language Models (VLMs) with Vision-Language-Action (VLA) models. Unlike prior VLM-based systems (e.g., VoxPoser, ReKep) that rely on depth perception and struggle with transparent labware, and existing VLA systems (e.g., RDT, $\pi_0$) that lack semantic-level feedback for complex tasks, our method leverages a VLM to serve as (1) a planner to decompose tasks into primitive actions, (2) a visual prompt generator to guide VLA models, and (3) a monitor to assess task success and regulatory compliance. Notably, we introduce a VLA interface that accepts image-based visual targets from the VLM, enabling precise, goal-conditioned control. Our system successfully executes both primitive actions and complete multi-step chemistry protocols. Results show \textbf{23.57\%} higher average success rate and a \textbf{0.298} average increase in compliance rate over state-of-the-art VLA baselines, while also demonstrating strong generalization to objects and tasks. 
   
\end{abstract}


\vspace{-0.2cm}
\begin{figure}[ht]
    \includegraphics[scale=0.285]{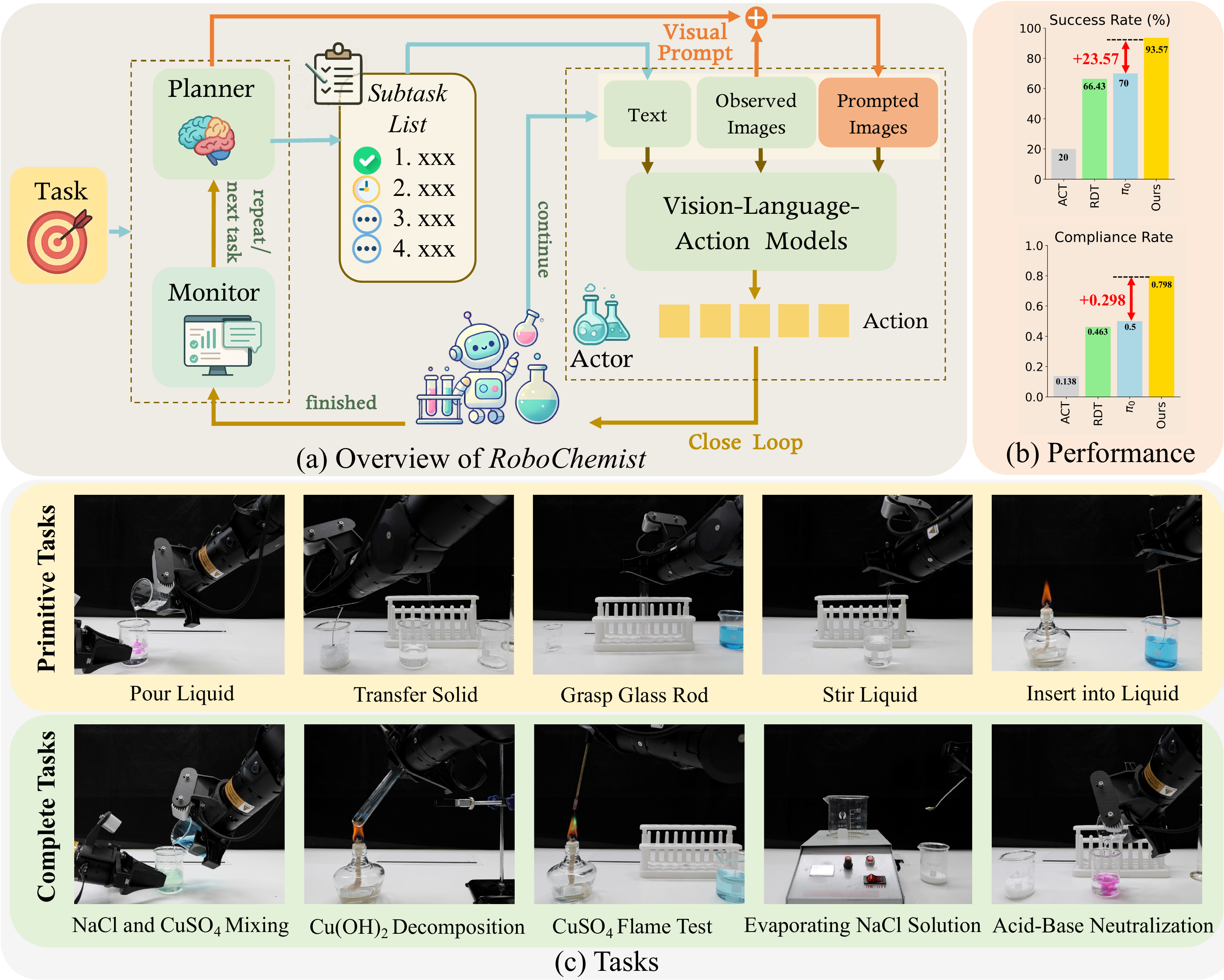} 
    \caption{(a) \textbf{Overview of \textit{RoboChemist}}. The VLM in our system acts as the planner, decomposing high-level tasks into subtasks. Based on each subtask, the VLM generates prompted images through visual prompting and provides them, along with other relevant information, to the VLA models. The VLM also functions as the monitor, assessing the completion status of subtasks, thus ensuring a complete feedback loop in the system. (b) \textit{RoboChemist} outperforms baselines in both primitive tasks and complete chemical experiment tasks. (c) Some tasks performed by \textit{RoboChemist}.
    }
     \vspace{-0.5cm}
    \label{fig:teaser}
\end{figure}

\section{Introduction}

Robotic chemists offer the potential to liberate human experts from repetitive and hazardous laboratory procedures, while accelerating scientific discovery~\citep{2020mobile_robotic_nature2020,material_synthesis_nature_2023,mobile_robots_synthetic_nature2024,organic_synthesis_auto_chemical_LLM_nature2023,science_2019organic_programming,chemical_synthesis_science2020,organic_synthesis_AI_planning_science2019}. Yet, the automation of chemical experiments remains highly challenging due to the long-horizon, safety-critical nature of lab protocols, which involve deformable substances, delicate glassware, and hazardous materials. These tasks require not only task success but also strict compliance with procedural norms—such as safe grasping points or heating thresholds—to ensure experimental validity and safety. As shown in Figure~\ref{fig:teaser}(c), even seemingly simple actions like stirring or pouring become complex when compositional reasoning, dexterous execution, and regulatory adherence are simultaneously required.

Recent progress in Vision-Language Models (VLMs)~\citep{llava, llava1.5, Internvl, cambrian1, qwen2.5vl,liu2023delving,li2022toist,jin2024tod3cap,chi2025impromptu,ding2024hint} and Vision-Language-Action (VLA) models~\citep{Openvla,octo_2023,RDT2025,pi_0,fast2025_pi0,pi0.5} has enabled scalable robotic policies across diverse environments. However, their limitations become pronounced in chemical labs. VLM-based systems like VoxPoser~\citep{voxposer2023} and ReKep~\citep{Rekep2024} rely heavily on depth sensors and object segmentation, which struggle with transparent containers and deformable substances~\citep{transparent_grasp_2023}. On the other hand, VLA models such as $\pi_0$~\citep{pi_0} and RDT~\citep{RDT2025} excel at action grounding but lack high-level semantic understanding or closed-loop feedback, often resulting in unsafe or failed behaviors in complex tasks. As illustrated in Figure~\ref{fig:teaser}(b), such approaches yield low success and compliance rates in primitive lab tasks.

To address these challenges, we propose \textit{RoboChemist}, a dual-loop framework that integrates VLMs and VLAs via visual prompting and semantic supervision. As illustrated in Figure~\ref{fig:teaser}(a), \textit{RoboChemist} employs a VLM as: (1) \textbf{a planner} that decomposes long-horizon tasks into structured subtasks; (2) \textbf{a visual prompt generator} that highlights task-specific grasp or target regions via bounding boxes or keypoints; and (3) \textbf{a monitor} that evaluates subtask success and enforces closed-loop corrections. The VLA model executes each primitive task using the prompted image, observed state, and textual instruction, enabling goal-conditioned and safety-compliant control. This design allows the system to reason globally while acting locally, combining high-level semantic structure with low-level dexterity.

We evaluate \textit{RoboChemist} across a wide spectrum of chemical tasks, including both primitive operations (e.g., pouring, stirring, grasping) and multi-step complete experiments (e.g., acid-base neutralization, flame tests)(Figure~\ref{fig:teaser}(c)). As shown in Figure~\ref{fig:teaser}(b), \textit{RoboChemist} achieves significant gains over prior baselines, with a \textbf{23.57\%} higher average success rate and a \textbf{0.298} average increase in compliance rate. The benefits of the closed-loop architecture are particularly evident in long-horizon tasks, where the outer loop enables condition-based re-execution (e.g., pouring until the solution becomes colorless). Additionally, our system generalizes to unseen reagents, containers, and workflows, demonstrating robust skill transfer without task-specific fine-tuning. These capabilities are further illustrated in Figure~\ref{fig:general tasks}, showcasing diverse experimental scenarios handled effectively by \textit{RoboChemist}.

To summarize, we make three contributions in this paper: (1) We introduce a dual-loop framework that unifies vision-language reasoning and low-level action grounding, with the VLM acting as planner, prompt generator, and monitor. (2) We propose an instruction-aware visual prompting strategy using Qwen2.5-VL, enabling precise and safe manipulation of transparent and hazardous materials in complex chemical setups. (3) \textit{RoboChemist} outperforms strong VLA and VLM baselines in both atomic and long-horizon tasks, demonstrating superior success, compliance, and generalization.


\vspace{-9pt}
\section{Related Work}
\vspace{-6pt}
\label{sec:related work}

\textbf{Vision-Language-Action Models.}
Large Language Models (LLMs)~\citep{gpt3, gpt4, llama,llama2, qwen2.5, deepseekllm} and Vision-Language Models (VLMs)~\citep{llava, llava1.5, Internvl, cambrian1, qwen2.5vl, zhang2025chameleon} have enabled scalable robot foundation models~\citep{RT-2, Openvla, octo_2023}. With robot learning datasets expanding from a few thousand~\citep{roboturk2018, bridgedata2021} to around one million samples~\citep{openx-embodiment2024, Droid2024}, and tasks evolving from single-arm~\citep{RT12022} to dual-arm~\citep{RDT2025,Agibot2025}, Vision-Language-Action (VLA) architectures are now being applied to embodied tasks~\citep{vl-foundation_robots_imitators, An_Embodied_Generalist_Agent_in_3D_World, interactive_agent_foundation_model,Tracevla2024, universal_foundation_model_air2025, towards_vla2024, Tinyvla2025, momanipvla_CVPR2025, jiang2025diffvla}. Recent VLA models differ in both perception and control design. While many use 2D visual inputs~\citep{pi_0, Cogact2024, spatialvla2025, ding2024preafford, Robomamba2024}, others incorporate 3D scene representations for richer spatial understanding~\citep{3DVLA2024, gr00TN1_nvidia2025}. On the action side, methods span simple MLP regressors~\citep{pi_0, Cogact2024, Robomamba2024}, autoregressive policy token decoders~\citep{Agibot2025, spatialvla2025, gemini_robotics2025}, and diffusion-based policies~\citep{RDT2025}. Hybrid architecture~\cite{hybridvla2025} and token-efficient designs~\citep{fast2025_pi0} further aim to balance performance and scalability.

\textbf{Visual Prompting.}
Visual prompting~\citep{visualprompting2022,visualprompttuning2022,promptinggpt4v2023,visualprompt2023,explicitvisualprompting2023,visualprompt2024,vipllavaprompting2024,visualprompt2025} has emerged as an effective strategy for guiding vision-language models by embedding task-specific visual cues into inputs. Recent visual prompting methods in robotics extract keypoints from RGB images via object masks, often using segmentation models~\citep{segmentanything,dinov2,clip,ViT} followed by Farthest Point Sampling (FPS)~\citep{FPS2003} or K-means~\citep{kmeans1999}, as in ReKep~\citep{Rekep2024} and KUDA~\citep{KUDA2025}. MOKA~\citep{moka} further integrates VLMs to select keypoints from candidates post-hoc. However, these pipelines lack direct use of textual prompts during keypoint generation, limiting adherence to instruction-level constraints—especially important in chemistry. In contrast, our method leverages advanced VLMs like Qwen2.5-VL~\citep{qwen2.5vl} to produce visual prompts that are instruction-aware from the outset.

\textbf{Robotic Automation in Chemistry.}
Robotic automation in chemistry has advanced significantly, covering organic synthesis~\citep{organic_synthesis_auto_chemical_LLM_nature2023,organic_synthesis_AI_planning_science2019}, material synthesis~\citep{material_synthesis_nature_2023}, microplate handling~\citep{evaluation_microplate_handling_accuracy_roboticarm2024}, and mobile exploratory chemists~\citep{2020mobile_robotic_nature2020,chemistry_auto_constrain_motion_planning2022,mobile_robots_synthetic_nature2024}. However, these systems typically depend on specialized hardware and predefined instructions, restricting flexibility and generalization across tasks~\citep{science_2019organic_programming,chemical_synthesis_science2020}. Recent efforts, including Organa~\citep{Organa2025}, ArChemist~\citep{ARchemist}, and liquid handling approaches~\citep{Pour_water2022,Pour_water2025}, partially address these limitations but remain predominantly single-arm and scenario-specific. The Chemistry3D benchmark~\citep{chemistry3d2024_simulation} introduces a simulated environment to evaluate general-purpose manipulation in chemistry contexts, failing to address the complexities of real-world chemical experimentation.

In contrast, our work introduces a generalized dual-arm collaborative approach utilizing foundational VLA models. This enables versatile manipulation across diverse chemical experiments without reliance on specialized hardware or rigid task-specific programming, aiming to improve practicality and scalability in robotic laboratory applications.

\vspace{-9pt}
\section{Method}
\vspace{-6pt}
\label{sec:method}

In this section, we first present an overview of the \textit{RoboChemist} pipeline in Sec.~\ref{sub:overview}. Sec.~\ref{sub: visual prompting} covers the motivation and approach using visual prompting, while Sec.~\ref{sub:Close loop} focuses on the design and implementation of the closed-loop system.

\begin{figure}[h]
  \centering
   \includegraphics[scale=1.0]{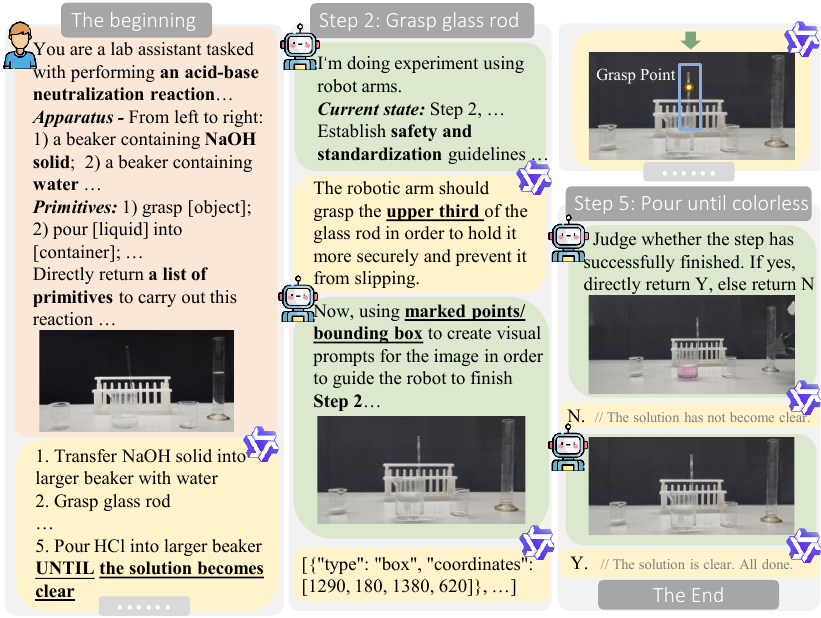}
   \caption{\textbf{Illustration of the acid-base neutralization reaction experiment.} 1) \textbf{The Beginning}: The researcher provides the complete task instructions, initial scene setup and available primitive tasks, followed by \textit{RoboChemist}'s task decomposition. 2) \textbf{Step 2}: During the glass rod grasping step, \textit{RoboChemist} uses visual prompting to highlight the \textcolor{blue}{bbox} and \textcolor[RGB]{192,144,0}{grasp point} to adhere to safety guidelines, providing reference for subsequent actions. 3) \textbf{Step 5}: \textit{RoboChemist}'s closed-loop system observes the experiment’s state and continues adding acid until the solution turns colorless, at which point the task is completed and the experiment is terminated.
}
   \label{fig: overview example}
   \vspace{-0.5cm}
\end{figure}

\vspace{-3pt}
\subsection{Overview}
\vspace{-3pt}
\label{sub:overview}
Figure~\ref{fig:teaser}(a) illustrates the overall pipeline of \textit{RoboChemist}, consisting of a closed-loop system formed by the VLM and VLA model. When the researcher specifies a chemical experiment task along with the apparatus and reagents, the VLM acts as a planner, decomposing the task into a sequence of executable primitive tasks. For each subtask, considering the specific operation protocols of chemical experiments, the VLM generates detailed guidelines, then uses visual prompting to highlight grasp points, target points, or bounding boxes in the front view (as shown in Figure~\ref{fig: overview example}). This reference image, along with observations from other camera viewpoints and the corresponding language instructions, is provided to the VLA model to execute each primitive task. After each subtask, the VLM functions as a monitor, evaluating the current state. If the subtask is completed successfully, the system proceeds to the next step; otherwise, the current task is repeated until successful completion. This forms a complete closed-loop feedback system, ensuring both safety and improved success rates, with high interpretability.

\vspace{-3pt}
\subsection{Visual Prompting}
\vspace{-3pt}
\label{sub: visual prompting}

\begin{wrapfigure}{r}{0.4\textwidth}
  \centering
  
  \vspace{-0.5cm}
  \begin{minipage}{0.39\textwidth}
  
  \resizebox{1\linewidth}{!}{
  \begin{tabular}{c|cc}
         \toprule
         Success Rate& Two Cups&Three Cups\\
         \midrule
         w/o visual prompt& 16/20 & 5/20\\
         \rowcolor{gray!20}w/ visual prompt& \textbf{19}/20 & \textbf{17}/20\\
         \bottomrule
    \end{tabular}
}
  \captionof{table}{Comparison of different methods for pouring liquid between cups.}
  \label{tab: motivation1}
  
  
 \vspace{-0.3cm}
  \end{minipage}
\end{wrapfigure}

\paragraph{Motivation.}  While VLA models exhibit strong capabilities in perceiving scenes, their abilities remain limited when conditioned solely on text prompts as instructions~\citep{zhaovlas, Tracevla2024}, especially in environment with many visually similar objects, like chemical experiments involving multiple containers. In such cases, visual prompting can largely alleviate the issue.
Here, we compare the performance of standalone VLA model with our visual-prompt-based method, \textit{RoboChemist}, on a liquid pouring task, which requires transferring liquid between two specific cups among a set of two or three.
Results in Table~\ref{tab: motivation1} indicate that performance degrades as the number of cups increases, underscoring the necessity of visual prompting.

\begin{wrapfigure}{r}{0.4\textwidth}
  \centering
  
  \vspace{-1.2cm}
  \begin{minipage}{0.39\textwidth}
  
  \includegraphics[width=1\textwidth]{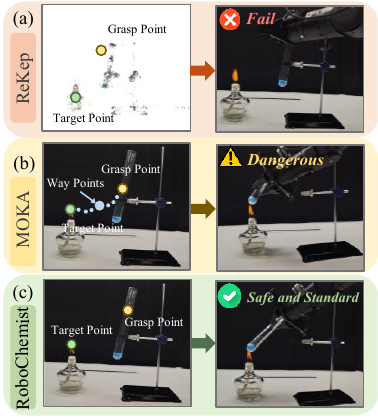}
  
  \captionof{figure}{Comparison of different methods for grasping a test tube containing solid for heating over a flame.}
  
 \vspace{-0.5cm}
  \label{fig:visual prompting}
  \end{minipage}
\end{wrapfigure}

However, while there are several visual prompt-based approaches, most of them face challenges in reconstructing transparent objects or ensuring safe, controlled execution~\citep{transparent_grasp_2023}. To illustrate this, we designed an experiment where a robotic arm is instructed to heat solid over a flame, as shown in Figure~\ref{fig:visual prompting}. 
In this experiment, ReKep~\citep{Rekep2024} struggles to reconstruct the transparent test tube, leading to a failed grasp. Methods that pre-compute grasp candidates without considering textual instructions are generally effective in standard settings but may fail to account for safety and procedural requirements crucial in chemistry experiments. As shown in Figure~\ref{fig:visual prompting}(b), MOKA~\citep{moka} selects the center as the grasp point, resulting in an unsafe action.

\paragraph{Proposed Method.} To overcome the limitations of existing methods, we introduce a framework to self-generate task-specific safety guidelines and visual prompts.
The method starts by providing the current state of the experiment—including the specific primitive task to be performed and images of the lab bench—as input to a VLM to generate detailed safety guidelines. 
These generated guidelines include considerations such as 
safe grasping regions on transparent labware, or chemical handling precautions—all without human intervention. 
Next, the model is tasked with using these contexts to produce visual prompts in the form of bounding boxes or key points. Thanks to the fine-grained grounding ability of Qwen-2.5-VL~\citep{qwen2.5vl}, we are able to reliably identify critical points/regions—even in visually ambiguous contexts. 

The proposed method demonstrates its end-to-end simplicity with high precision. As illustrated in Figure~\ref{fig:visual prompting}(c), the grasp point is placed on the upper body of the test tube—avoiding the heated area—while the target point ensures correct positioning for material heating. A full example of the visual prompt generation process is illustrated in Figure~\ref{fig: overview example}, where in Step 2, the robotic arm is tasked with grasping a glass rod.
By bridging the gap between perception, instruction generation, and actionable spatial grounding, we establish a new paradigm for safe, instruction-aware robotics in chemically sensitive environments.

\vspace{-3pt}
\subsection{Closed-Loop System Design}
\vspace{-3pt}

\label{sub:Close loop}
\textbf{Inner Loop Enhancement with Unsuccessful Trials.} The VLA models naturally have an inner closed loop, obtaining real-time images and state information as inputs and generating actions to interfere the environment. However, the loop would end after a failed attempt. To enhance the self-correcting ability and robustness of the loop, we introduce scenarios into the training data where, after an unsuccessful trial, the system automatically attempts a second execution. This approach, rather than relying solely on successful one-shot tasks, improves the VLA model’s feedback capabilities, resulting in a stronger inner loop.

\textbf{Outer Loop Establishment with Monitors.} The inner loop lacks an explicit feedback mechanism, making it unable to distinguish between success and failure or support long-horizon behaviors that require ongoing monitoring. \textit{RoboChemist} introduces an outer loop integrating the VLM as a monitor. After each inner loop, the current scene images are fed to the VLM, which evaluates whether each primitive task has achieved its intended goal and enables recovery from failures. For complex, multi-repetition tasks, the outer loop can also provide ongoing feedback. For instance, in Step 9 of Figure~\ref{fig: overview example}, the initial attempt to pour acid does not fully neutralize the base—evidenced because the phenolphthalein indicator remaining pink. Therefore, the outer loop triggers a re-execution of the pouring primitive until the solution clear. Through this mechanism, the outer loop composes multiple discrete actions into a coherent sequence that approximates continuous behavior.

Additionally, the outer loop functions as a planner, decomposing complex tasks into sequences of primitive subtasks. As shown at the beginning of Figure~\ref{fig: overview example}, \textit{RoboChemist} uses the known apparatus configuration to break down an acid-base neutralization reaction into a series of primitives.



\vspace{-9pt}
\section{Experiment}
\vspace{-6pt}
\label{sec:result}

\subsection{Experiment Setup}
\vspace{-3pt}
\textbf{Hardware Platform.} The system uses the Cobot Magic ALOHA, a dual-arm robot with 7 degrees of freedom per arm. It is equipped with four Depth Camera D435 units, but we only utilize RGB data from cameras on the two wrists, front, and a top-down viewpoint. An NVIDIA RTX 4090 GPU handles data collection and model inference.

\textbf{Data Collection.} For VLA model fine-tuning, we collect 400 data samples for each primitive task; the full list of primitive tasks is provided in Sec.~\ref{sec:chemical tasks}. To ensure task generalization, we introduce diversity in both the experimental scene setup and object selection for each basic action. For example, in the \textit{liquid pouring} task, we vary the liquid color (e.g., blue, red, colorless, etc.), volume (e.g., small, medium, large, etc.), container type (e.g., beakers, test tubes, graduated cylinders, etc.), and the surrounding environment layout. The data collection frequency aligns with the inference frequency, both set at 20 Hz. All data is stored in Parquet or HDF5 format for training.

\textbf{Baselines.} To evaluate \textit{RoboChemist}, we consider advanced baselines in robotic bimanual manipulation: ACT~\citep{ACT}, RDT~\citep{RDT2025}, and $\pi_0$~\citep{pi_0}. All models are fine-tuned using our collected data under the same setup. ACT is a state-of-the-art method in bimanual manipulation, utilizing action chunking with transformers. RDT-1B uses the DiT architecture, the latest dual-arm foundation model. $\pi_0$, using the VLA architecture, demonstrates strong generalization across tasks and robotic platforms.

\textbf{Metric.} We evaluate each primitive task using two metrics: \textbf{Success Rate} (SR) and \textbf{Compliance Rate} (CR).
For each task, we conduct 20 trials. SR is calculated as the ratio of successful trials to total trials. To assess the precision of actions in chemical experiments, we introduce CR. For example, in the \textit{Grasp Glass Rod} task, a failed grasp is scored as 0, a grasp at an incorrect position (e.g., not at the 1/3 point of the rod) is scored as 0.5, and a successful and compliant grasp is scored as 1. Detailed CR settings for additional tasks can be found in the Appendix~\ref{app:primitive-tasks}.

\textbf{Model Training and Inference.} We fine-tune the $\pi_0$~\citep{pi_0} model with collected data, training each task for 30K steps on four L20 GPUs. During training, in addition to the images from the four camera viewpoints used for data collection, we apply Qwen2.5-VL-72B-Instruct~\citep{qwen2.5vl} visual prompting to label the grasp and target points in each primitive task’s initial experimental scene. These images, along with the language instructions and the proprioceptive state, are used as input to fine-tune the VLA model. To diversify the language instructions, we generate various command variations using GPT-4o~\citep{gpt4o}. Inference is performed in real-time on an NVIDIA RTX 4090 GPU at 20 Hz. 

\vspace{-3pt}
\subsection{Chemical Tasks}
\vspace{-3pt}
\label{sec:chemical tasks}
\textbf{Primitive Tasks.} To enable a wider range of chemical experiments, we decompose standard chemical tasks into 7 primitive tasks, including grasping a glass rod, heating a platinum wire, inserting platinum wire into a solution, pouring liquid, stirring a solution with a glass rod, transferring the solid, and pressing a button. For a more detailed illustration, please refer to the Appendix~\ref{app:primitive-tasks}. Table~\ref{tab:primitive tasks} compares \textit{RoboChemist} and baselines on these primitive tasks in terms of Success Rate (SR) and Compliance Rate (CR), with all models fine-tuned using the same data. \textit{RoboChemist} outperforms all baselines in both SR and CR across the tasks. \textit{RoboChemist} without the closed-loop system (only the addition of a visual prompting reference image) shows improvement compared to $\pi_0$~\citep{pi_0}, demonstrating the effectiveness of visual prompting. Furthermore, \textit{RoboChemist} with the outer-loop structure achieves higher success rates than the version without the outer loop, highlighting the effectiveness of the closed-loop system in enhancing primitive task completion.

\begin{table*}[ht]
  \centering
  \setlength{\tabcolsep}{3pt}
  \resizebox{1\linewidth}{!}{
  \begin{tabular}{l|cc|cc|cc|cc|cc|cc|cc}
    \toprule
    Primitive Task & \multicolumn{2}{c|}{Grasp Glass Rod} & \multicolumn{2}{c|}{Heat Platinum Wire} & \multicolumn{2}{c|}{Insert into Solution} & \multicolumn{2}{c|}{Pour Liquid} & \multicolumn{2}{c|}{Stir the Solution} & \multicolumn{2}{c|}{Transfer the Solid} & \multicolumn{2}{c}{Press the Button}\\
    \midrule
    Method & SR(\%) $\uparrow$ & CR $\uparrow$ & SR(\%) $\uparrow$ & CR $\uparrow$ & SR(\%) $\uparrow$ & CR $\uparrow$ & SR(\%) $\uparrow$ & CR $\uparrow$ & SR(\%) $\uparrow$ & CR $\uparrow$ & SR(\%) $\uparrow$ & CR $\uparrow$ & SR(\%) $\uparrow$ & CR $\uparrow$\\
    \midrule
    ACT~\citep{ACT} & 55 & 0.325 & 20  & 0.063 & 10  & 0.050 & 25  & 0.288 & 15  & 0.075 & 15  & 0.063 & 0  & 0.100\\
    RDT~\citep{RDT2025} & 20  & 0.100   & 60  & 0.363 & 80  & 0.775  & 90  & 0.675 & 75  & 0.400 & 75  & 0.513 & 65  & 0.413\\
    $\pi_0$~\citep{pi_0} & 40  & 0.200  & 55  & 0.325 & 80  & 0.800  & 80  & 0.475  & 85  & 0.600 & 80  & 0.525 & 70  & 0.575\\ 
    \rowcolor{gray!20}
    RoboChemist w/o CL & 85  & 0.750  & 70  & 0.575 & 85  & 0.850  & 80  & 0.663  & 95  & 0.650 & 85  & 0.538 & 75  & 0.613\\
    \rowcolor{gray!20} 
    RoboChemist w/ CL & \textbf{95}   & \textbf{0.875}   & \textbf{90}   & \textbf{0.800} & \textbf{95}   & \textbf{0.950}   & \textbf{95}   & \textbf{0.800} & \textbf{100}   & \textbf{0.825} & \textbf{95}   & \textbf{0.675} & \textbf{85}   & \textbf{0.663}\\
    \bottomrule
  \end{tabular}
  }
  \caption{\textbf{Primitive tasks results.} We evaluate baselines and \textit{RoboChemist} on seven primitive tasks using Success Rate (SR) and average Compliance Rate (CR). CL denotes Closed-Loop architecture.}
  \label{tab:primitive tasks}
  \vspace{-0.5cm}
\end{table*}

\begin{figure}[h]
  \centering
   \includegraphics[width=0.9\linewidth]{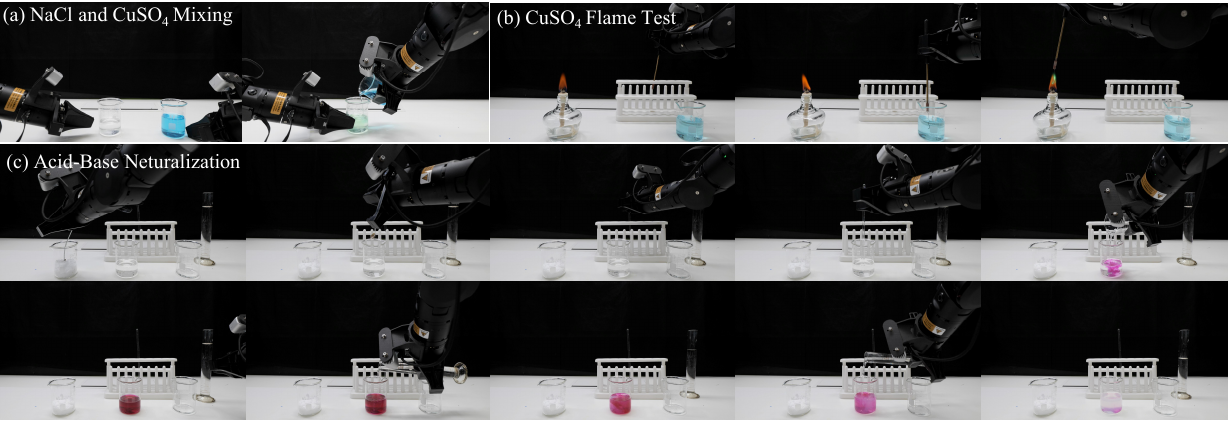}
   \caption{\textbf{Visualization of Complete Chemical Experiments.} (a) Complexation reaction: mixing NaCl and CuSO\textsubscript{4} to form a light green complex; (b) Copper flame test: dipping a platinum wire into CuSO\textsubscript{4} solution and observing a green flame upon heating; (c) Acid-base neutralization: preparing a NaOH solution and using phenolphthalein as an indicator to determine neutralization by HCl.
}
   \label{fig: Qualitative results Complete tasks}
\end{figure}

\textbf{Complete Tasks.} After training the primitive tasks, we proceed to execute complete experimental tasks. We select five complete chemical experiment tasks (detailed description please refer to Appendix~\ref{app:complete-tasks}),  where each task consists of 1 to 5 primitive tasks in order. Figure~\ref{fig: Qualitative results Complete tasks} shows the experimental process and phenomena as \textit{RoboChemist} completes the tasks. Table~\ref{tab:complete tasks} shows the performance metrics for each primitive task in order. As the experiments progress, other baselines experience a decline in success rate; however, \textit{RoboChemist} performs consistently well in tasks composed of 2-3 primitive tasks, and even successfully completes more complex tasks involving up to five steps. For unseen tasks, such as grasping a test tube, \textit{RoboChemist} also demonstrates strong generalization. 

\vspace{-0.3cm}

\begin{table*}
  \centering
  \setlength{\tabcolsep}{3pt}
  \resizebox{1\linewidth}{!}{
  \begin{tabular}{l|cc|cc|cc|cc|cc|cc|cc}
    \toprule
    Complete Task & \multicolumn{2}{c|}{Mix NaCl and Cu(SO$_4$)} & \multicolumn{4}{c|}{Heat-induced Decompose Cu(OH)$_2$} & \multicolumn{8}{c}{Flame Test for CuSO$_4$ Solution}\\
    \midrule
    Primitive Task & \multicolumn{2}{c|}{Pour Liquid} & \multicolumn{2}{c|}{Grasp Test Tube} & \multicolumn{2}{c|}{Heat Test Tube} & \multicolumn{2}{c|}{Grasp Platinum Wire} &\multicolumn{2}{c|}{Heat Platinum Wire} & \multicolumn{2}{c|}{Insert into Solution} & \multicolumn{2}{c}{Heat Platinum Wire}\\
    \midrule
    Method & SR(\%) $\uparrow$ & CR $\uparrow$ & SR(\%) $\uparrow$ & CR $\uparrow$ & SR(\%) $\uparrow$ & CR $\uparrow$ & SR(\%) $\uparrow$ & CR $\uparrow$ & SR(\%) $\uparrow$ & CR $\uparrow$ & SR(\%) $\uparrow$ & CR $\uparrow$ & SR(\%) $\uparrow$ & CR $\uparrow$\\
    \midrule
    ACT~\citep{ACT}& 25  & 0.250   & 45  & 0.250 & 30  & 0.150  & 50  & 0.500 & 10  & 0.062 & 10  & 0.100 & 0  & 0.000\\
    RDT~\citep{RDT2025}& 80  & 0.650   & 10  & 0.050 & 5  & 0.025  & 20  & 0.200 & 10  & 0.075 & 10  & 0.100 & 5  & 0.038\\
    $\pi_0$~\citep{pi_0}& 80  & 0.650   & 40  & 0.250 & 35  & 0.200  & 40  & 0.400 & 25  & 0.200 & 20  & 0.200 & 15  & 0.125\\ 
    \rowcolor{gray!20}
    RoboChemist  & \textbf{95}  & \textbf{0.775}   & \textbf{90}  & \textbf{0.900} & \textbf{80}  & \textbf{0.450}  & \textbf{90}  & \textbf{0.900} & \textbf{80}  & \textbf{0.750} & \textbf{80}  & \textbf{0.800} & \textbf{65}  & \textbf{0.650}\\
    \bottomrule
        Complete Task & \multicolumn{4}{c|}{Evaporate NaCl Solution} & \multicolumn{10}{c}{Acid-Base Neutralization Reaction}\\
    \midrule
    Primitive Task & \multicolumn{2}{c|}{Transfer NaCl Solid} & \multicolumn{2}{c|}{Press the Button} & \multicolumn{2}{c|}{Transfer NaOH Solid} & \multicolumn{2}{c|}{Grasp Glass Rod} &\multicolumn{2}{c|}{Stir the Solution} & \multicolumn{2}{c|}{Add Phenolphthalein} & \multicolumn{2}{c}{Add HCl Acid}\\
    \midrule
    Method & SR(\%) $\uparrow$ & CR $\uparrow$ & SR(\%) $\uparrow$ & CR $\uparrow$ & SR(\%) $\uparrow$ & CR $\uparrow$ & SR(\%) $\uparrow$ & CR $\uparrow$ & SR(\%) $\uparrow$ & CR $\uparrow$ & SR(\%) $\uparrow$ & CR $\uparrow$ & SR(\%) $\uparrow$ & CR $\uparrow$\\
    \midrule
    ACT~\citep{ACT} & 15  & 0.063   & 0  & 0.100 & 20  & 0.075  & 10  & 0.050 & 5  & 0.025 & 0  & 0.000 & 0  & 0.000\\
    RDT~\citep{RDT2025} & 75  & 0.513   & 50  & 0.363 & 75  & 0.513  & 20  & 0.100 & 15  & 0.075 & 5  & 0.038 & 0  & 0.013\\
    $\pi_0$~\citep{pi_0} & 80  & 0.525   & 55  & 0.400 & 80  & 0.525  & 35  & 0.200 & 30  & 0.175 & 20  & 0.150 & 5  & 0.013\\ 
    \rowcolor{gray!20}
    RoboChemist & \textbf{95}  & \textbf{0.675}   & \textbf{80}  & \textbf{0.600} & \textbf{100}  & \textbf{0.675}  & \textbf{95}  & \textbf{0.850} & \textbf{90} & \textbf{0.650} & \textbf{80}  & \textbf{0.625} & \textbf{40}  & \textbf{0.300}\\
    \bottomrule
  \end{tabular}
  }
  \caption{\textbf{Complete tasks results.} We evaluate baselines and \textit{RoboChemist} on five complete chemical experiments using Success Rate (SR) and Compliance Rate (CR). For tasks that require multiple steps, the metrics for each subsequent step are calculated based on the success of the previous step.}
  \label{tab:complete tasks}
  \vspace{-0.3cm}

\end{table*}

\vspace{-0.5cm}
\subsection{Effectiveness of Visual Prompting}
\vspace{-5pt}

To evaluate the effectiveness of our proposed visual prompting method, we compare it with the $\pi_0$~\citep{pi_0} baseline and other visual prompt-based methods, including ReKep~\citep{Rekep2024} and MOKA~\citep{moka}. In all experiments, we generate visual prompts from the same input instructions and use the resulting images as references for the fine-tuned VLA model. Table~\ref{tab:visual prompting} presents the quantitative results. In the case of ReKep~\citep{Rekep2024}, the RGB-D cameras struggle with the reconstruction of transparent objects, leading to the lowest success rates. MOKA~\citep{moka}, not relying on text-based visual prompts, shows relatively lower compliance. In contrast, \textit{RoboChemist}, using a VLM-based visual prompt, achieves the best results, demonstrating the superiority of our method in effectively utilizing visual cues.

\begin{table*}
  \centering
  \setlength{\tabcolsep}{3pt}
  \resizebox{1\linewidth}{!}{
  \begin{tabular}{l|cc|cc|cc|cc|cc|cc|cc}
    \toprule
   Primitive Task & \multicolumn{2}{c|}{Grasp Glass Rod} & \multicolumn{2}{c|}{Heat Platinum Wire} & \multicolumn{2}{c|}{Insert into Solution} & \multicolumn{2}{c|}{Pour Liquid} & \multicolumn{2}{c|}{Stir the Solution} & \multicolumn{2}{c|}{Transfer the Solid} & \multicolumn{2}{c}{Press the Button}\\
    \midrule
    Method & SR(\%) $\uparrow$ & CR $\uparrow$ & SR(\%) $\uparrow$ & CR $\uparrow$ & SR(\%) $\uparrow$ & CR $\uparrow$ & SR(\%) $\uparrow$ & CR $\uparrow$ & SR(\%) $\uparrow$ & CR $\uparrow$ & SR(\%) $\uparrow$ & CR $\uparrow$ & SR(\%) $\uparrow$ & CR $\uparrow$\\
    \midrule
    $\pi_0$~\citep{pi_0} & 40  & 0.200  & 55  & 0.325 & 80  & 0.800  & \textbf{80}  & 0.475  & 85  & 0.600 & 80  & 0.525 & 70  & 0.363\\
    ReKep~\citep{Rekep2024} + $\pi_0$~\citep{pi_0} & 35  & 0.200   & 30  & 0.150 & \textbf{85}  & 0.825  & 55  & 0.375 & 90  & 0.500 & 75  & 0.313 & \textbf{75}  & 0.400\\
    MOKA~\citep{moka} + $\pi_0$~\citep{pi_0} & 65  & 0.350   & 55  & 0.338 & \textbf{85}  & \textbf{0.850}  & 75  & 0.500 & \textbf{95}  & 0.600 & \textbf{85}  & 0.363 & \textbf{75}  & 0.475\\ 
    \rowcolor{gray!20} 
    RoboChemist w/o CL & \textbf{85}   & \textbf{0.750}   & \textbf{70}   & \textbf{0.575} & \textbf{85}   & \textbf{0.850}    & \textbf{80}  & \textbf{0.663}  & \textbf{95}  & \textbf{0.650} & \textbf{85}  & \textbf{0.538} & \textbf{75}  & \textbf{0.613}\\
    \bottomrule
  \end{tabular}
  }
  \caption{Comparison of visual prompt-based methods.}
  \vspace{-0.3cm}
  \label{tab:visual prompting}
\end{table*}

\vspace{-0.1cm}
\subsection{Generalizability}
\vspace{-5pt}

\begin{table*}
  \centering
  \setlength{\tabcolsep}{8pt}  
  \renewcommand{\arraystretch}{1.2}  
  \resizebox{\textwidth}{!}{
  \begin{tabular}{l|c|c}
    \hline
    \textbf{Primitive Task} & \textbf{Training Data} & \textbf{Generalized Tasks} \\
    \hline
    Pick-and-Place & Grasp a glass rod & Grasp and place test tubes, beakers, graduated cylinders, etc \\
    \hline
    Stir & Stir the solution with a glass rod & Stir solid reagents or use different objects to stir \\
    \hline
    Heat & Heat a platinum wire & Heat various types of materials, such as iron wire, magnesium wire, test tubes, etc \\
    \hline
    Pour liquid & Pour liquid from one container to another & Measure liquid solvents\\
    \hline
    Insert & Insert platinum wire into a solution & Insert any solid reagent, a thermometer, or place a test tube into cooling liquid.\\
    \hline
  \end{tabular}
  }
  \caption{Primitive Tasks Generalization.}
  \label{tab:primitive tasks generalization}
\end{table*}

Leveraging the VLM's open vocabulary capability~\citep{vlm-openvocabulary} and the VLA's generalization ability, \textit{RoboChemist} demonstrates remarkable adaptability, successfully performing both primitive and complete tasks across a wide range of scenarios. Please refer to Appendix~\ref{sec:app_generalization} for more details. 

\textbf{Generalization in Primitive Tasks.} As shown in Table~\ref{tab:primitive tasks generalization}, \textit{RoboChemist} generalizes its learned primitive tasks to various objects and materials. For instance, grasping a glass rod was trained and successfully extends to picking and placing test tubes, beakers, and other containers. These examples highlight its robust capability to apply basic skills to diverse experimental scenarios.

\begin{figure}[h]
  \centering
   \includegraphics[width=1.0\linewidth]{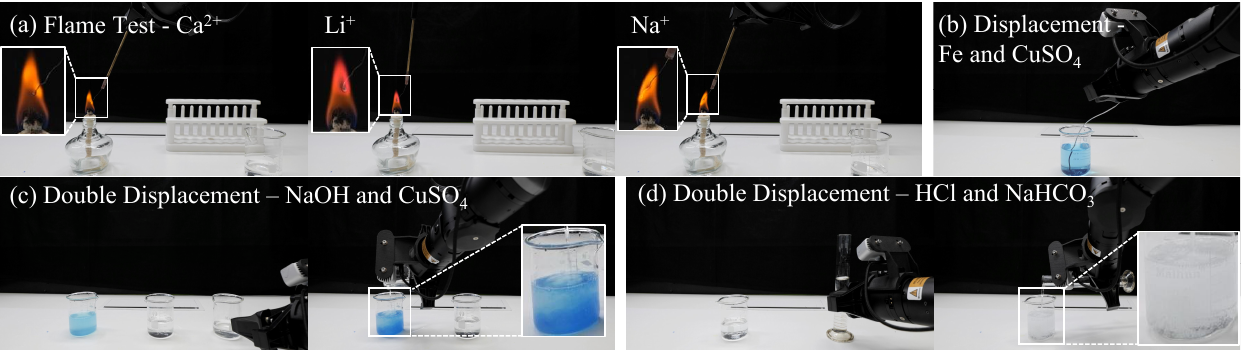}
   \caption{\textbf{Visualization of Generalization in Complete Tasks.} (a) Flame tests: Ca\textsuperscript{2+} (brick-red), Li\textsuperscript{+} (purplish-red), and Na\textsuperscript{+} (yellow); (b) Displacement reaction: Fe displaces Cu from CuSO\textsubscript{4} solution; (c) Double displacement reaction: NaOH and CuSO\textsubscript{4} form Cu(OH)\textsubscript{2} precipitate; (d) Double displacement reaction: HCl and NaHCO\textsubscript{3} generate CO\textsubscript{2} gas bubbles.
    }
   \label{fig:general tasks}
\end{figure}

\textbf{Generalization in Complete Tasks.} 
Although only seven primitive tasks were trained, \textit{RoboChemist} leverages their generalization and the extensive chemical knowledge stored in the VLM to perform a wide range of complete tasks.  This ability not only allows it to complete the specified tasks but also enables it to infer and summarize experimental observations, drawing conclusions based on the outcomes. Figure~\ref{fig:general tasks} demonstrates how \textit{RoboChemist} can execute the four basic types of chemical reactions—combination, decomposition, displacement, and double displacement reactions. It also explores the physical and chemical properties of different metal elements, such as flame reactions and comparing the reactivity of metals through displacement reactions with solutions.

\vspace{-9pt}
\section{Conclusion}
\vspace{-5pt}
\label{sec:conclusion}
In this paper, we present \textit{RoboChemist}, a dual-loop framework combining Vision-Language Models (VLMs) with Vision-Language-Action (VLA) models for safe and interpretable robotic chemical experiments. The VLM serves as a planner, visual prompt generator, and monitor—decomposing protocols, guiding VLA models, and verifying outcomes with corrective feedback. This closed-loop design handles hazardous and complex labware while ensuring safety and procedural compliance. Evaluations on primitive actions and multi-step tasks show \textit{RoboChemist} surpasses prior methods in success and compliance. It generalizes well to new reagents, containers, and workflows, highlighting its potential to enhance lab automation with less human input and greater reliability.


\section{Limitations}
\label{sec:limitation}
While \textit{RoboChemist} demonstrates strong capabilities, several limitations remain. First, the gripper used for execution is not specifically designed for chemical tasks, limiting its ability to handle certain instruments; incorporating a more specialized robotic hand would expand the range of tasks that can be performed. Second, the system is currently unable to autonomously assemble chemical experimental setups, which require high precision and careful handling of various components. Tasks like connecting delicate glassware or positioning sensitive equipment demand more intricate manipulation capabilities that go beyond the current robotic system's design. Achieving this would require the integration of more advanced manipulation techniques and finer control over the robot’s movements. Lastly, \textit{RoboChemist}'s current architecture is not suited for tasks requiring strict quantification and precise time control. Future work includes integrating more capable hardware, extending the model to support fine-grained temporal and quantitative control, and testing in real-world laboratory settings.

\bibliography{references}  
\newpage

\setcounter{page}{1}

\begin{center}
    {\LARGE \textbf{RoboChemist: Long-Horizon and Safety-Compliant}}\\
    {\LARGE \textbf{Robotic Chemical Experimentation}}
\end{center}

\begin{appendices}
\section{Appendix}

In this appendix, we provide comprehensive supplementary materials and detailed technical insights to support and extend the results presented in the main paper. Sec.~\ref{app:primitive-tasks} details the primitive tasks, including specific compliance criteria and evaluation metrics used to measure success rate and compliance rate. Sec.~\ref{app:complete-tasks} describes the five complete chemical experiments used in our evaluation,elaborating on chemical principles, experimental protocols, and observation criteria for multi-step tasks. Sec.~\ref{sec:app_keyprompts} lists key prompts utilized for task decomposition, safety guideline generation, visual prompting, and task completion verification. Sec.~\ref{sec:app_generalization} presents additional evaluation demonstrating RoboChemist’s capability for generalization across primitive tasks and complete chemical procedures. Sec.~\ref{sec:app_visual_prompt_methods} compares various visual prompting strategies to assess their effectiveness in guiding robotic manipulations. Sec.~\ref{sec:app-inner-loop} quantitatively evaluates the impact of training data configurations with varied proportions of successful trials and second-attempt exploratory trials on the robustness and self-correcting capabilities of the inner loop. Sec.~\ref{sec:app-vla-architecture} describes the specific VLA architecture employed in our work. Sec.~\ref{sec:app-error-breakdown}  provides an error decomposition across five modules. Finally, Sec.~\ref{sec:app-gen-env-embodiment} evaluates the generalization of one task under diverse variations, showing that VLM feedback and visual prompting enable robust transfer.

\subsection{Details of Primitive Tasks, Compliance Criteria, and Evaluation Protocols}
\label{app:primitive-tasks}
This section provides a detailed description of the seven primitive tasks performed by \textit{RoboChemist}, including the task objectives and the compliance criteria used for evaluation. The performance of each primitive task is evaluated using two metrics: Success Rate (SR) and Compliance Rate (CR), where SR represents the ratio of successful trials to total trials, and CR measures how well the task adheres to predefined experimental norms.

\begin{figure}[h]
  \centering
   \includegraphics[width=1.0\linewidth]{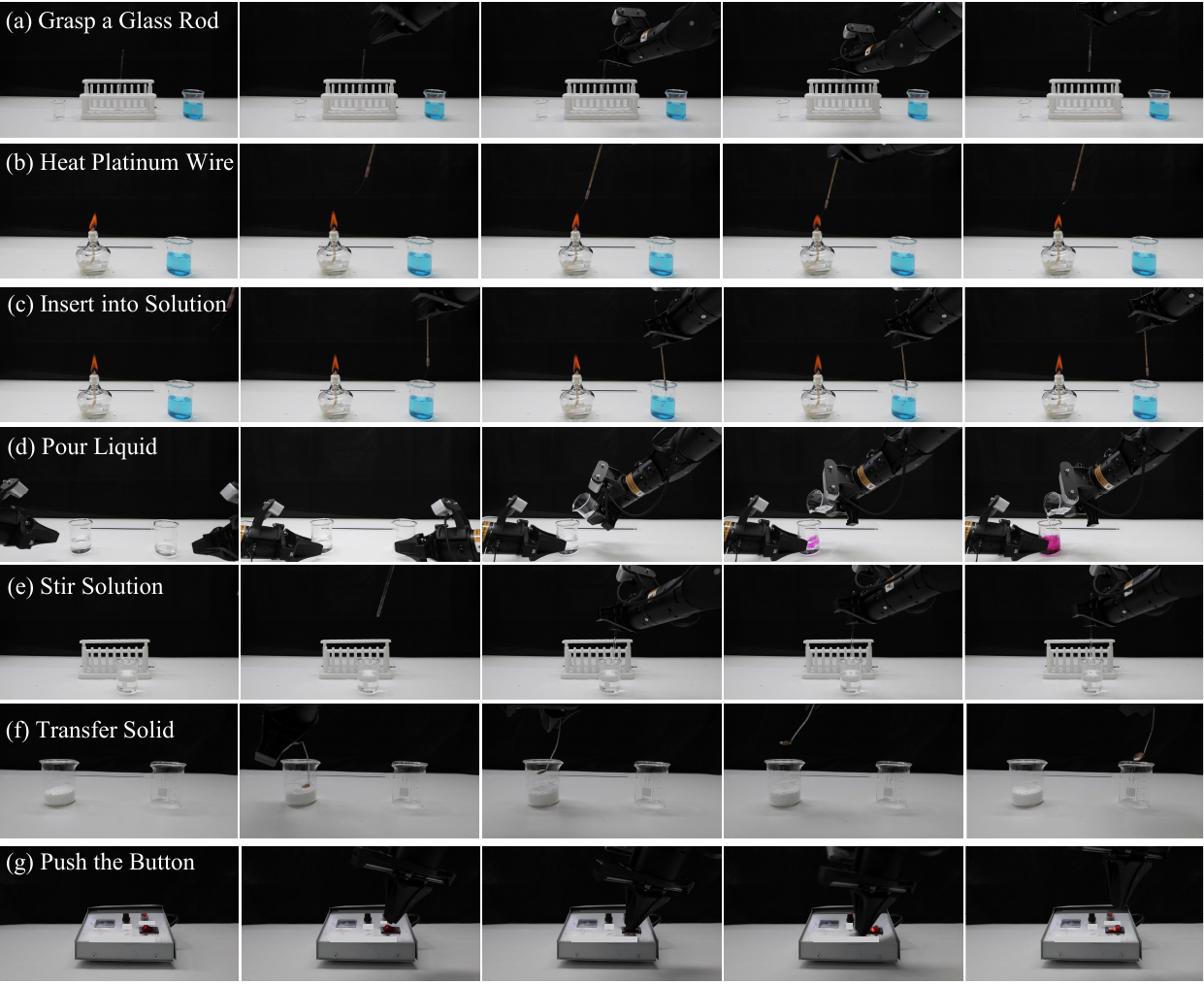}
   \caption{Visualization of primitive tasks.
}
   \label{fig:app_primitive}
\end{figure}

\begin{enumerate}
    \item \textbf{Grasping a Glass Rod:}
    \begin{itemize}
        \item Objective: Grasp a glass rod successfully.
        \item Compliance Criteria:
        \begin{itemize}
            \item Grasp fails (does not grasp the rod): 0
            \item Grasp is made but not at the correct position (e.g., not at the 1/3 point of the rod): 0.5
            \item Correct grasp at the 1/3 position of the rod: 1
        \end{itemize}
    \end{itemize}

    \item \textbf{Heating Platinum Wire:}
    \begin{itemize}
        \item Objective: Heat a platinum wire in a flame to a specific region.
        \item Compliance Criteria:
        \begin{itemize}
            \item Does not heat: 0
            \item Heats at the wrong location and does not return: 0.25
            \item Heats at the wrong location but returns: 0.5
            \item Heats at the correct location but does not return: 0.75
            \item Heats at the correct location and returns: 1
        \end{itemize}
    \end{itemize}

    \item \textbf{Inserting Platinum Wire into Solution:}
    \begin{itemize}
        \item Objective: Insert a platinum wire into a solution.
        \item Compliance Criteria:
        \begin{itemize}
            \item Does not reach the top of the beaker: 0
            \item Reaches the beaker but does not insert into the solution: 0.5
            \item Correctly inserts into the solution: 1
        \end{itemize}
    \end{itemize}

    \item \textbf{Pouring Liquid:}
    \begin{itemize}
        \item Objective: Pour a liquid from one container into another.
        \item Compliance Criteria:
        \begin{itemize}
            \item Fails to grasp: 0
            \item Grasped but spills completely: 0.25
            \item Grasped but spills slightly: 0.75
            \item Successfully grasps and pours liquid with control: 1
        \end{itemize}
    \end{itemize}

    \item \textbf{Stirring Solution with a Glass Rod:}
    \begin{itemize}
        \item Objective: Stir a solution to ensure proper mixing.
        \item Compliance Criteria:
        \begin{itemize}
            \item Does not contact liquid: 0
            \item Stirred but uneven, hitting the container wall: 0.5
            \item Correct stirring with adequate speed and uniform direction: 1
        \end{itemize}
    \end{itemize}

    \item \textbf{Transferring Solid:}
    \begin{itemize}
        \item Objective: Transfer a solid from one container to another.
        \item Compliance Criteria:
        \begin{itemize}
            \item Does not transfer solid: 0
            \item Solid transferred but spills out: 0.25
            \item Solid transferred with slight spillage: 0.75
            \item Solid transferred without spillage: 1
        \end{itemize}
    \end{itemize}

    \item \textbf{Pressing a Button:}
    \begin{itemize}
        \item Objective: Press a button to initiate a process.
        \item Compliance Criteria:
        \begin{itemize}
            \item Does not press: 0
            \item Pressed but not activated: 0.25
            \item Pressed but too forcefully, causing movement of the equipment: 0.5
            \item Correct press with appropriate force: 1
        \end{itemize}
    \end{itemize}
\end{enumerate}

Based on this, the Success Rate (SR) is calculated by performing the task 20 times and dividing the number of successful trials by the total number of trials. Specifically, the Success Rate is given by:
\[
SR = \frac{\text{Number of successful trials}}{\text{Total number of trials}}.
\]
The Compliance Rate (CR) is the average score of the 20 trials, where each trial is assigned a score based on the compliance with the task criteria. For example, in the "\textit{Grasping a Glass Rod}" task, suppose there are \(n_1\) trials where the score is 0, \(n_2\) trials with a score of 0.5, and \(n_3\) trials with a score of 1. Then the Compliance Rate is calculated as:

\[
CR = \frac{(0 \times n_1) + (0.5 \times n_2) + (1 \times n_3)}{n_1 + n_2 + n_3}.
\]

In this example, \(n_1 + n_2 + n_3 = 20\). The CR is the average of the scores obtained in these 20 trials, representing how well the actions align with the experimental norms.

This calculation method is used to derive the data presented in Table~\ref{tab:primitive tasks} for the primitive tasks, where both Success Rate and Compliance Rate are computed for each task. \textbf{Note:} The last two rows in Table~\ref{tab:primitive tasks} correspond to different variants of \textit{RoboChemist}. \textit{RoboChemist w/o CL} represents the version without the outer closed-loop; that is, the VLM is not used to assess whether each primitive task is successfully completed, while the visual prompting module is still included. In contrast, \textit{RoboChemist w/ CL} denotes the complete system. In this configuration, the VLM acts as a monitor: after each primitive execution, it evaluates the current scene and determines whether the task has succeeded. If not, it instructs the VLA model to reattempt the primitive until the task is deemed successful, forming a full closed-loop feedback mechanism that enhances both robustness and compliance. By comparing the performance of these two variants, we observe the significant role of the closed-loop system in improving the success and compliance of primitive task execution.

\subsection{Details of Complete Chemical Tasks and Evaluation Protocols}
\label{app:complete-tasks}
This section provides detailed descriptions of the five complete chemical experiment tasks used in our evaluation. Each task is composed of 1 to 5 primitive tasks arranged sequentially, and is designed to cover a diverse range of chemical operations and reaction types. For each task, we describe the objective, underlying reaction principle, expected observable phenomena, and the language prompts used to initiate the sequence.

\textbf{1. Mixing NaCl and CuSO\textsubscript{4} Solutions}  

\begin{figure}[h]
  \centering
   \includegraphics[width=1.0\linewidth]{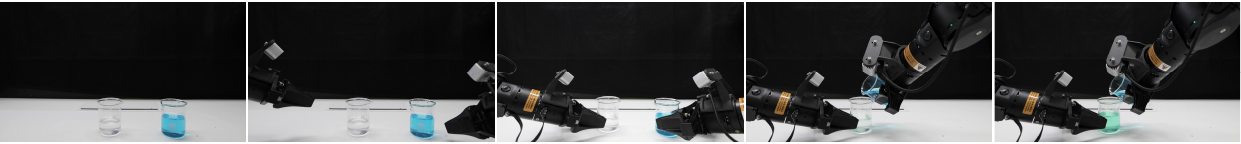}
   \caption{Visualization of mixing NaCl and CuSO\textsubscript{4} solutions.
}
   \label{fig:app_complete1}
\end{figure}

\begin{itemize}
    \item \textbf{Objective:} Mix sodium chloride solution with copper sulfate solution and observe the resulting color change.
    \item \textbf{Primitive tasks:} Pour one beaker of liquid into another.
    \item \textbf{Observation and Explanation} This is a coordination reaction in which hydrated copper ions, initially blue in color, react with chloride ions to form the complex ion [CuCl\textsubscript{4}]\textsuperscript{2-}. The reaction proceeds as:
    \[
    \mathrm{[Cu(H_2O)_4]^{2+} + 4Cl^- \rightarrow [CuCl_4]^{2-} + 4H_2O}
    \]
    As the reaction occurs, the solution color transitions from blue to cyan and then to green. The final green hue corresponds to the formation of sodium tetrachlorocuprate, Na\textsubscript{2}[CuCl\textsubscript{4}], indicating successful complexation (as Figure~\ref{fig:app_complete1} shows).

\end{itemize}

\textbf{2. Thermal Decomposition of Cu(OH)\textsubscript{2}}  

\begin{figure}[h]
  \centering
   \includegraphics[width=1.0\linewidth]{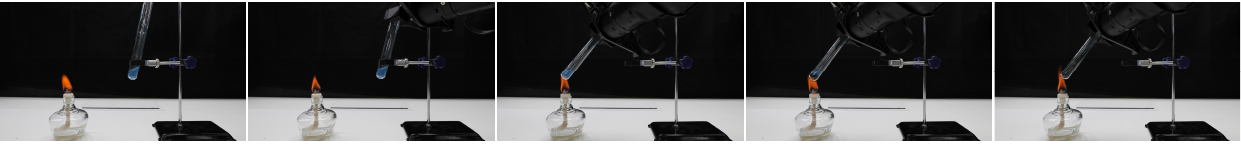}
   \caption{Visualization of thermal decomposition of Cu(OH)\textsubscript{2}.
}
   \label{fig:app_complete2}
\end{figure}

\begin{itemize}
    \item \textbf{Objective:} Heat a test tube containing solid Cu(OH)\textsubscript{2} and observe the resulting decomposition process.
    
    \item \textbf{Primitive tasks:} Grasp the test tube containing Cu(OH)\textsubscript{2} $\rightarrow$ Heat over flame.
    
    \item \textbf{Explanation and Observation:} Copper(II) hydroxide (Cu(OH)\textsubscript{2}) is a pale blue solid. Upon heating, it decomposes into copper(II) oxide (CuO), which is black, and water vapor. During the reaction, mist forms on the inner wall of the test tube due to condensation of steam. The chemical equation is:
    \[
    \mathrm{Cu(OH)_2 \xrightarrow{\Delta} CuO + H_2O}
    \]
    This reaction reflects the thermal instability of Cu(OH)\textsubscript{2} and the stability of CuO at elevated temperatures (as Figure~\ref{fig:app_complete2} shows).
    
\end{itemize}

\textbf{3. Flame Test of CuSO\textsubscript{4} Solution}  

\begin{figure}[h]
  \centering
   \includegraphics[width=1.0\linewidth]{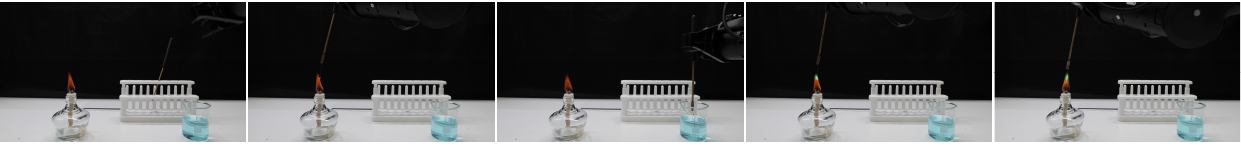}
   \caption{Visualization of flame test of CuSO\textsubscript{4} solution.
}
   \label{fig:app_complete3}
\end{figure}

\begin{itemize}
    \item \textbf{Objective:} Identify the presence of Cu\textsuperscript{2+} ions in copper sulfate through a flame test, investigating its physical property via characteristic flame emission.
    
    \item \textbf{Primitive tasks:} Grasp platinum wire $\rightarrow$ Dip into CuSO\textsubscript{4} solution $\rightarrow$ Heat platinum wire in flame.
    
    \item \textbf{Observation and Explanation:} When a compound containing Cu\textsuperscript{2+} ions (e.g., CuSO\textsubscript{4}) is heated in a flame, it emits a characteristic blue-green color. This flame color originates from the emission spectrum of Cu\textsuperscript{2+} ions, which release photons primarily in the 450–500 nm wavelength range when thermally excited. The resulting blue-green light is a reliable indicator of the presence of copper ions and is commonly used in qualitative elemental analysis (as Figure~\ref{fig:app_complete3} shows).
    
\end{itemize}

\textbf{4. Evaporation of NaCl Solution}  

\begin{figure}[h]
  \centering
   \includegraphics[width=1.0\linewidth]{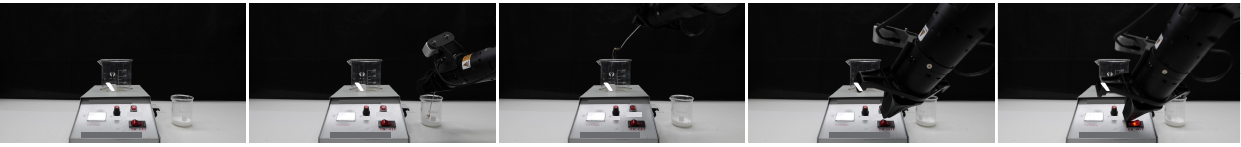}
   \caption{Visualization of evaporation of NaCl solution.
}
   \label{fig:app_complete4}
\end{figure}

\begin{itemize}
    \item \textbf{Objective:} Evaporate an impure NaCl solution to separate soluble salt from insoluble impurities.
    
    \item \textbf{Primitive tasks:} Transfer solid NaCl into a beaker of water $\rightarrow$ Press the heater button to initiate evaporation.
    
    \item \textbf{Explanation and Observation:} The primary component of the crude salt is sodium chloride (NaCl), which dissolves readily in water to form a solution, while insoluble impurities such as sand or other salts remain at the bottom. Upon pressing the heater button, the solution begins to heat and eventually boils, producing visible bubbles. As heating continues, the water gradually evaporates, increasing the salt concentration. Once the solution reaches saturation, solid NaCl crystals begin to precipitate. When all water has evaporated, white NaCl crystals remain in the beaker, typically hard in texture and free from soluble contaminants (as Figure~\ref{fig:app_complete4} shows).
    
\end{itemize}

\textbf{5. Acid-Base Neutralization with Phenolphthalein Indicator}  

\begin{figure}[h]
  \centering
   \includegraphics[width=1.0\linewidth]{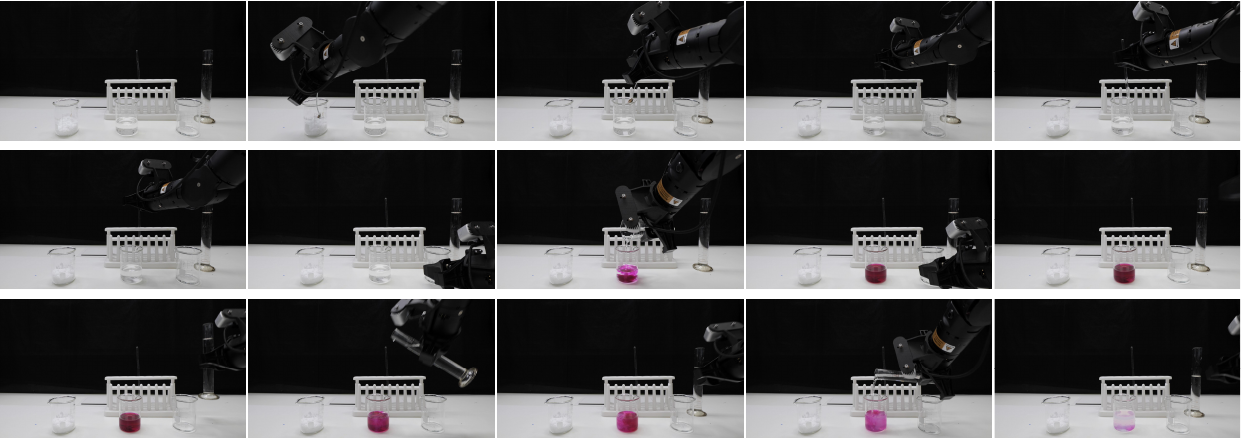}
   \caption{Visualization of acid-base neutralization with phenolphthalein indicator.
}
   \label{fig:app_complete5}
\end{figure}

\begin{itemize}
    \item \textbf{Objective:} Neutralize a sodium hydroxide (NaOH) solution by gradually adding hydrochloric acid (HCl) until the solution reaches neutrality, as indicated by phenolphthalein.
    
    \item \textbf{Primitive tasks:} Transfer solid NaOH into a beaker of water $\rightarrow$ Grasp a glass rod $\rightarrow$ Stir the NaOH solution to dissolve the solid $\rightarrow$ Add phenolphthalein indicator $\rightarrow$ Pour HCl into the NaOH solution until the solution becomes colorless.

     \item \textbf{Explanation and Observation:} Sodium hydroxide is a strong base and dissolves in water to form a colorless alkaline solution. Phenolphthalein is added as an acid-base indicator, turning the solution red in basic conditions. As hydrochloric acid is gradually added, H\textsuperscript{+} ions react with OH\textsuperscript{-} ions to form water, reducing the pH of the solution. When neutrality is achieved, the pink color disappears and the solution becomes colorless (as Figure~\ref{fig:app_complete5} shows). This indicates the end point of the acid-base neutralization reaction:
    \[
    \mathrm{HCl + NaOH \rightarrow NaCl + H_2O}
    \]

\end{itemize}

Each experiment is prompted through natural language instructions, and \textit{RoboChemist} decomposes the tasks and monitors the progress via its dual-loop system. To quantitatively evaluate the execution of complete tasks, we follow a stepwise evaluation protocol aligned with the sequential structure of chemical experiments. As shown in Table~\ref{tab:complete tasks}, each complete task is composed of a series of primitive tasks. We conduct 20 independent trials for each complete task. If any primitive task within the sequence fails, all subsequent steps are considered unsuccessful for that trial. Therefore, the success rate (SR) for a complete task is computed as the proportion of trials in which all constituent primitive tasks are successfully completed. Similarly, the compliance rate (CR) for each primitive step is averaged only over the trials where that specific step was reached. Consequently, the final step’s SR in each row also reflects the overall SR of the complete task, serving as a comprehensive indicator of the whole task's success.


\subsection{Key Prompts}
\label{sec:app_keyprompts}
\subsubsection{Task Decomposition}
We present examples of our proposed prompt for task decomposition in each chemical experiment. For different chemical experiments, only the apparatus and task need to be modified.

\textbf{1. Mixing NaCl and CuSO\textsubscript{4} Solutions}  
\begin{tcolorbox}[enhanced, breakable, fontupper=\ttfamily]
You are a lab assistant tasked with mixing NaCl and CuSO$_4$ solutions to form sodium tetrachlorocuprate using the materials shown in the image. The items, from left to right, are:

\begin{itemize}
    \item A beaker with sodium chloride solution.
    \item A beaker with copper sulfate solution.

\end{itemize}
\par\ \par
\textbf{Task:} Mixing NaCl and CuSO$_4$ solutions to form sodium tetrachlorocuprate.
\par\ \par
\textbf{Goal:} Provide a step-by-step list of primitives (simple lab actions) required to carry out this reaction safely and correctly using the available materials.
\par\ \par
\textbf{Output Format:} Directly return a list of primitive actions only containing the following steps without any explanations. If given until [condition] sentence, the robot will repeat the operation until the condition is achieved:

\begin{itemize}
    \item Grasp [rod-like object]
    \item Pour [liquid] from [container] into [container] until [condition]
    \item Stir [mixture]
    \item Transfer [solid] from [container] to [container]
    \item Dip [object] into the [solution] in [container]
    \item Heat [object] over a flame
    \item Press the button of [object]
    
\end{itemize}
\end{tcolorbox}
\textbf{2. Thermal Decomposition of Cu(OH)\textsubscript{2}}  
\begin{tcolorbox}[enhanced, breakable, fontupper=\ttfamily]
You are a lab assistant tasked with performing the thermal decomposition of copper(II) hydroxide using the materials shown in the image. The items, from left to right, are:

\begin{itemize}
    \item A lit alcohol lamp.
    \item A test tube containing copper(II) hydroxide.

\end{itemize}
\par\ \par
\textbf{Task:} Performing the thermal decomposition of copper(II) hydroxide.
\par\ \par
\ldots
\end{tcolorbox}

\textbf{3. Flame Test of CuSO\textsubscript{4} Solution}  
\begin{tcolorbox}[enhanced, breakable, fontupper=\ttfamily]
You are a lab assistant tasked with performing the flame test of copper(II) hydroxide using the materials shown in the image. The items, from left to right, are:

\begin{itemize}
    \item A lit alcohol lamp.
    \item Platinum wire.
    \item A test tube containing copper(II) hydroxide.

\end{itemize}
\par\ \par
\textbf{Task:} Performing the flame test of copper(II) hydroxide.
\par\ \par
\ldots
\end{tcolorbox}

\textbf{4. Evaporation of NaCl Solution}  
\begin{tcolorbox}[enhanced, breakable, fontupper=\ttfamily]
You are a lab assistant tasked with evaporating an impure NaCl solution to separate soluble salt from insoluble impurities using the materials shown in the image. The items, from left to right, are:

\begin{itemize}
    \item An evaporator with a power button.
    \item A breaker with NaOH solid.

\end{itemize}
\par\ \par
\textbf{Task:} Evaporating an impure NaCl solution to separate soluble salt from insoluble impurities
\par\ \par
\ldots
\end{tcolorbox}

\textbf{5. Acid-Base Neutralization with Phenolphthalein Indicator}  
\begin{tcolorbox}[enhanced, breakable, fontupper=\ttfamily]
You are a lab assistant tasked with performing an acid-base neutrali- zation reaction using the materials shown in the image. The items, from left to right, are:

\begin{itemize}
    \item A beaker with NaOH solid.
    \item A beaker with water and a glass rod.
    \item A beaker with phenolphthalein indicator.
    \item A graduated cylinder containing hydrochloric acid (HCl).
    \item A glass rod in a test tube rack.

\end{itemize}
\par\ \par
\textbf{Task:} Performing an acid-base neutralization reaction.
\par\ \par
\ldots
\end{tcolorbox}

\subsubsection{Safety Guideline Generation} 
We present examples of our proposed prompt for safety guideline generation prior to performing primitive tasks. For different primitive tasks, only the name of each task needs to be modified.
\begin{tcolorbox}[enhanced, breakable, fontupper=\ttfamily]
    I am doing experiment using robot arms. Regarding to step [NUMBER], [PRIMITIVE], establish safety and standardization guidelines necessary for conducting the chemistry experiment for robot arms, such as where are the items related to this step, where grasp points of robot arm are, etc. These guidelines should be only related to items mentioned before. Summarize these in one or two clear sentences. If such guidelines are not applicable or available, return None. 
\end{tcolorbox}
\subsubsection{Visual Prompt Generation} 
We present examples of our proposed prompt for visual prompt generation prior to performing primitive tasks. For different primitive tasks, only the name of each task needs to be modified.
\begin{tcolorbox}[enhanced, breakable, fontupper=\ttfamily]
    
Now, given the image of the current state of experiment, using marked points/bounding box to create visual prompts for the image in order to guide the robot to finish step [NUMBER], [PRIMITIVE]. Remember that only mark the points/bounding box that robot can currently see/operate in the current state and do not predict future things/items. Return a list of point/bounding box coordinates in the image of current state in the format of a list, like [{"type": "box", "coordinates":[xmin, ymin, xmax, ymax]}, {"type":"point", "coordinates":[x, y]}]. If there is nothing to return, return an empty list.
\end{tcolorbox}

\subsubsection{Task Completion Verification} 

We present examples of our proposed prompt for task completion verification after performing primitive tasks. For different primitive tasks, only the name of each task needs to be modified.
\begin{tcolorbox}[enhanced, breakable, fontupper=\ttfamily]
Judge whether the step [NUMBER], [PRIMITIVE], has successfully finished from the image provided. If yes, directly return Y, else return N.

\end{tcolorbox}

\subsection{Extended Analysis of Generalization}
\label{sec:app_generalization}

\begin{figure}[h]
  \centering
   \includegraphics[width=1.0\linewidth]{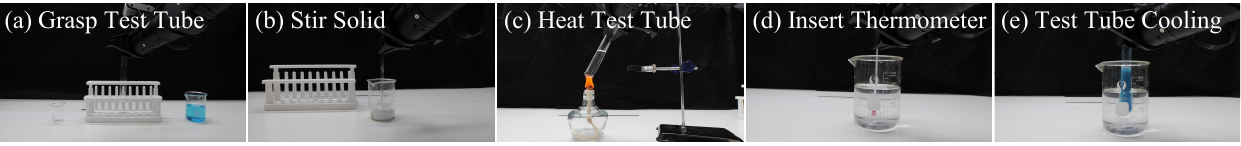}
   \caption{Visualization of primitive task generalization.
}
   \label{fig:app_primitive_general}
\end{figure}

\textbf{Primitive Task Generalization.} 
To further validate the generalization capability of \textit{RoboChemist} at the primitive task level, we design and evaluate a set of novel tasks that are not seen during training but are compositionally similar to the original primitives (as shown in Figure~\ref{fig:app_primitive_general}). These tasks introduce variations in object geometry and physical interaction while preserving the core manipulation intent. We conduct 20 trials for each task, using the same evaluation criteria as in the main paper—Success Rate (SR) and Compliance Rate (CR). As shown in Table~\ref{tab:primitive tasks generalization}, \textit{RoboChemist} demonstrates strong transferability, successfully applying learned manipulation policies to novel scenarios with consistent safety and execution quality.

\begin{table*}[ht]
  \centering
  \setlength{\tabcolsep}{3pt}
  \resizebox{1\linewidth}{!}{
  \begin{tabular}{l|cc|cc|cc|cc|cc|cc}
    \toprule
    Training Task & \multicolumn{4}{c|}{Grasp Glass Rod} & \multicolumn{2}{c|}{Stir the Solution} &  \multicolumn{2}{c|}{Heat Platinum Wire} & \multicolumn{4}{c}{Insert Platinum Wire into Solution}\\
    \midrule
    Generalized Task & \multicolumn{2}{c|}{Place Glass Rod} & \multicolumn{2}{c|}{Grasp Test Tube} & \multicolumn{2}{c|}{Stir Solid Reagents} & \multicolumn{2}{c|}{Heat Test Tube} & \multicolumn{2}{c|}{Insert a Thermometer} & \multicolumn{2}{c}{Place Test Tube into Cooling Liquid}\\
    \midrule
    Method & SR(\%) $\uparrow$ & CR $\uparrow$ & SR(\%) $\uparrow$ & CR $\uparrow$ & SR(\%) $\uparrow$ & CR $\uparrow$ & SR(\%) $\uparrow$ & CR $\uparrow$ & SR(\%) $\uparrow$ & CR $\uparrow$ & SR(\%) $\uparrow$ & CR \\
    \midrule
    ACT~\citep{ACT} & 10 & 0.050 & 40  & 0.250 & 15  & 0.075 & 5  & 0.025 & 10  & 0.025 & 10  & 0.050 \\
    RDT~\citep{RDT2025} & 0 & 0.000 & 10  & 0.050 & 60  & 0.350 & 35  & 0.175 & 75  & 0.500 & 60  & 0.300 \\
    $\pi_0$~\citep{pi_0} & 35 & 0.200 & 40  & 0.250 & 85  & 0.450 & 55  & 0.425 & 80  & 0.550 & 65  & 0.450 \\ 
    \rowcolor{gray!20}
    RoboChemist & \textbf{55} &\textbf{ 0.400} & \textbf{90}  & \textbf{0.850} & \textbf{100}  & \textbf{0.785} & \textbf{70}  & \textbf{0.675} & \textbf{95}  & \textbf{0.950} & \textbf{85}  & \textbf{0.850} \\
    \bottomrule
  \end{tabular}
  }
  \caption{Primitive task generalization results.}
  \label{tab:quantitative general primitive tasks}
\end{table*}

\textbf{Complete Task Generalization.} Beyond the generalization of individual primitives, \textit{RoboChemist} shows the ability to generalize to novel complete chemistry tasks composed of sequences of primitive actions. This is enabled by the VLM's capacity to reason over unseen goals and plan appropriate action sequences using its chemical knowledge. In particular, the system successfully performs complete experiments corresponding to the four fundamental types of chemical reactions: combination, decomposition, displacement, and double displacement.

\begin{figure}[h]
  \centering
   \includegraphics[width=1.0\linewidth]{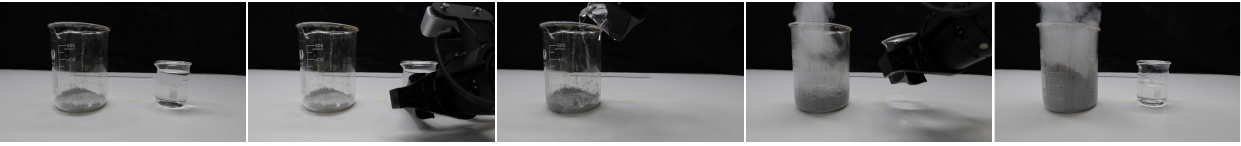}
   \caption{Visualization of combination reaction.
}
   \label{fig:app_complete_CaO}
\end{figure}

\textbf{(a) Combination Reaction:} \textit{RoboChemist} pours water into a container of calcium oxide (CaO), initiating a vigorous exothermic reaction that forms calcium hydroxide:
\[
\mathrm{CaO + H_2O \rightarrow Ca(OH)_2}
\]
During execution, the calcium oxide rapidly reacts with water, releasing a large amount of heat. This thermal energy generates a hissing sound and partial vaporization of the liquid. The reaction produces a white solid—calcium hydroxide—which partially dissolves, forming a milky suspension. This example demonstrates the system's ability to manage high-temperature reactions and recognize key visual and auditory indicators of successful chemical transformation (as Figure~\ref{fig:app_complete_CaO} shows).

\begin{figure}[h]
  \centering
   \includegraphics[width=1.0\linewidth]{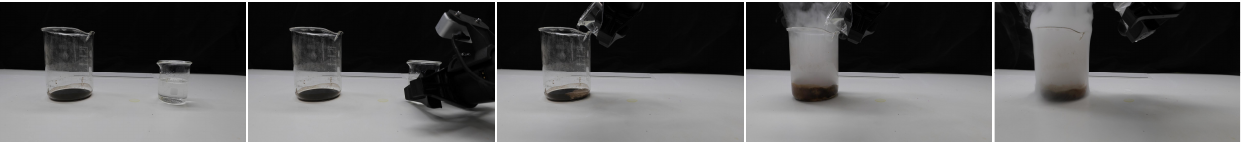}
   \caption{Visualization of decomposition reaction.
}
   \label{fig:app_complete_H2O2}
\end{figure}

\textbf{(b) Decomposition Reaction:} \textit{RoboChemist} performs a catalytic decomposition of hydrogen peroxide (H\textsubscript{2}O\textsubscript{2}) by introducing Manganese(II) hydroxide (Mn(OH)\textsubscript{2}) as a catalyst. The reaction proceeds as:
\[
2\text{H}_2\text{O}_2 \xrightarrow{\text{Mn(OH)}_{2}} 2\text{H}_2\text{O} + \text{O}_2 \uparrow
\]
Upon the addition of Mn(OH)\textsubscript{2}, \textit{RoboChemist} observes a rapid generation of oxygen gas, visible as dense bubbling at the liquid surface. This demonstrates the system’s ability to manage reactive liquid mixing, recognize real-time gas evolution, and safely interpret catalyst-driven decomposition reactions (as Figure~\ref{fig:app_complete_H2O2} shows).

\begin{figure}[h]
  \centering
   \includegraphics[width=1.0\linewidth]{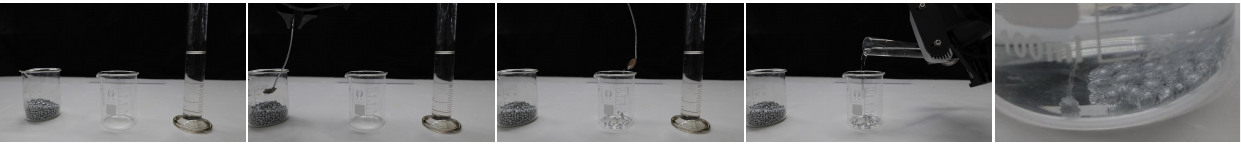}
   \caption{Visualization of displacement reaction between Zn and HCl.
}
   \label{fig:app_gen_complete_Zn}
\end{figure}

\textbf{(c) Displacement Reaction:} As Figure~\ref{fig:general tasks}(b) shows, \textit{RoboChemist} inserts an iron wire into a CuSO\textsubscript{4} solution. The reaction proceeds as:
\[
\text{Fe} + \text{CuSO}_4 \rightarrow \text{FeSO}_4 + \text{Cu}
\]
This causes reddish solid copper to precipitate on the iron surface, demonstrating iron's higher reactivity. Similarly, RoboChemist performs acid-metal reactions such as:
\[
\text{Zn} + 2\text{HCl} \rightarrow \text{ZnCl}_2 + \text{H}_2 \uparrow
\]
where bubbles form as hydrogen gas evolves and zinc dissolves, showcasing its ability to execute and interpret multiple forms of displacement processes (as Figure~\ref{fig:app_gen_complete_Zn} shows).

\textbf{(d) Double Displacement Reaction:} As Figure~\ref{fig:general tasks}(c)-(d) shows, \textit{RoboChemist} executes precipitation and gas-generation reactions through compound exchange. For example:
\[
2\text{NaOH} + \text{CuSO}_4 \rightarrow \text{Cu(OH)}_2 \downarrow + \text{Na}_2\text{SO}_4
\]
forming a distinct blue precipitate of Cu(OH)\textsubscript{2}. Another example includes:
\[
\text{NaHCO}_3 + \text{HCl} \rightarrow \text{NaCl} + \text{H}_2\text{O} + \text{CO}_2 \uparrow
\]
where CO\textsubscript{2} gas is produced, visible as vigorous bubbling, highlighting RoboChemist's capacity to identify evolving gases and reaction completion.

In addition, \textit{RoboChemist} can also distinguish between different metal ions based on flame color, revealing generalization to physical property inference. As shown in Figure~\ref{fig:general tasks}(a), RoboChemist performs flame tests by dipping a platinum wire into various solutions and observing the resulting flame color when heated. These colors serve as indicators of metal identity, including Ca\textsuperscript{2+} (brick-red), Li\textsuperscript{+} (purplish-red), Na\textsuperscript{+} (yellow), Mn\textsuperscript{2+} (yellow-green), and Sr\textsuperscript{2+} (magenta). 

This demonstrates that once \textit{RoboChemist} masters a representative instance of a given experimental category—such as a flame test—it can generalize this capability to a broader class of similar procedures. Such compositional generalization underscores the system’s ability to integrate procedural knowledge (via VLM) and low-level execution (via VLA) to generalize from basic actions to diverse, multi-step chemistry tasks. The system not only completes complex workflows but also interprets and responds to observed phenomena, showcasing its robust reasoning and adaptability. Please refer to the \href{https://zzongzheng0918.github.io/RoboChemist.github.io/}{project page} for qualitative demonstrations of generalization performance.

\subsection{Comparison of Visual Prompting Methods}
\label{sec:app_visual_prompt_methods}

This section provides detailed descriptions of the visual prompting strategies used in our comparison experiments (as shown in Table~\ref{tab:visual prompting}), specifically focusing on ReKep~\citep{Rekep2024}, MOKA~\citep{moka}, and our proposed VLM-based approach. All three methods generate visual prompts that are used as auxiliary supervision for fine-tuning a $\pi_0$~\citep{pi_0} policy model. This standardized setup enables a fair evaluation of the prompt generation quality across different frameworks.

In \textbf{ReKep}, visual prompts are created in the form of Relational Keypoint Constraints (ReKep). These constraints encode desired spatial relations between semantically meaningful 3D keypoints extracted from RGB-D input. ReKep first uses Large Vision Models (e.g., DINOv2) to generate dense keypoint candidates, then employs a Vision-Language Model (e.g., GPT-4o) to generate Python functions expressing constraints over these keypoints. For example, the constraint might require one keypoint (e.g., spout) to align vertically above another (e.g., cup center), or define a 3D vector between two points to encode a specific rotation. Although ReKep can flexibly describe complex spatio-temporal tasks, its performance suffers in scenes involving transparent or textureless objects, where depth-based keypoint proposals become unreliable. Additionally, since the constraints are encoded as code, they must still be transformed into explicit labeled examples (e.g., target positions) to be used for training $\pi_0$.

\textbf{MOKA}, on the other hand, formulates visual prompting as a mark-based problem. It uses open-vocabulary detectors (e.g., Grounding DINO) and segmentation models (e.g., SAM) to locate objects of interest, and then leverages Vision-Language Models (e.g., GPT-4V) to generate mark sets on RGB images — each mark corresponding to a 2D location (or region) semantically tied to the instruction. These marked images are treated as visual prompts to supervise a visuomotor agent. Importantly, MOKA avoids explicit 3D reconstruction and instead operates directly in 2D pixel space. While this allows for flexibility and ease of deployment, it lacks fine-grained geometric reasoning and results in lower task compliance, especially for cases requiring precise spatial alignment or multi-step reasoning.

In contrast, our method uses a vision-language model to directly annotate 2D visual prompts by pointing to semantically and geometrically relevant locations on input images. These annotated keypoints are automatically derived from the instruction-image pair using a VLM with built-in grounding ability. Unlike ReKep or MOKA, our approach does not rely on RGB-D input, depth reconstruction, or open-vocabulary object detectors, and produces supervision that is both semantically aligned and geometrically grounded. We then use these VLM-derived visual prompts as reference targets to fine-tune a VLA model, which can then serve as a policy for $\pi_0$ training.

Quantitative results in Table~\ref{tab:visual prompting} highlight that ReKep achieves limited success due to the difficulty of keypoint localization in challenging scenes (e.g., transparent objects), and MOKA exhibits moderate compliance as it lacks explicit spatial reasoning. In contrast, \textit{RoboChemist}, powered by VLM-based 2D prompting, consistently outperforms these methods, demonstrating the advantage of our direct, image-based visual prompting pipeline.

\subsection{Inner Loop Enhancement: Implementation and Evaluation}
\label{sec:app-inner-loop}

To investigate the impact of mixed success data on the performance of our VLA model, we conduct a controlled study using four distinct training configurations for each primitive task. Each configuration consists of a total of 400 training samples for a given task, but the proportion of successful trials and those with a second attempt after an initial failure varies:

\begin{itemize}
    \item \textbf{Config 1 (400/0 - All Successful):} 400 successful trials with no failed attempts.
    \item \textbf{Config 2 (300/100 - Majority Successful):} 300 successful trials and 100 trials where a second attempt is made after an initial failure.
    \item \textbf{Config 3 (200/200 - Balanced):} 200 successful trials and 200 trials where a second attempt is made after an initial failure.
    \item \textbf{Config 4 (0/400 - All Second Attempt):} 400 trials where a second attempt is made after an initial failure.
\end{itemize}

\begin{table*}[ht]
  \centering
  \setlength{\tabcolsep}{3pt}
  \resizebox{1\linewidth}{!}{
  \begin{tabular}{l|c|c|c|c|c|c|c|c}
    \toprule
    Primitive Task & \multicolumn{1}{c|}{Grasp Glass Rod} & \multicolumn{1}{c|}{Heat Platinum Wire} & \multicolumn{1}{c|}{Insert into Solution} & \multicolumn{1}{c|}{Pour Liquid} & \multicolumn{1}{c|}{Stir the Solution} & \multicolumn{1}{c|}{Transfer the Solid} & \multicolumn{1}{c|}{Press the Button} & \multicolumn{1}{c}{Average}\\
    \midrule
   
    Config 1 & \textbf{40} & 50 & 75  & \textbf{85} & \textbf{85}  & \textbf{80} & 55 & 67.14\\
    Config 2 & \textbf{40} & \textbf{55} & \textbf{80}  & 80 & \textbf{85} & \textbf{80} & 70 & \textbf{70}\\
    Config 3 & 35 & \textbf{55} & 70  & 65 & 65  & 70 & \textbf{75} & 62.14\\ 
    Config 4 & 25 & 20 & 15  & 5 & 30  & 15 & 45 & 22.14\\
    \bottomrule
  \end{tabular}
  }
  \caption{Impact of Different Training Configurations on Primitive Task Success Rates.}
  \label{tab:inner loop enhancement}
\end{table*}

We fine-tune the $\pi_0$ model using each of these four data configurations and quantitatively evaluate their impact on the success rate of the seven primitive tasks. Table~\ref{tab:inner loop enhancement} presents the success rates for each task under the four configurations. The results demonstrate that Config 2 (300/100) achieves the highest average success rate (70\%), indicating the optimal balance between successful and exploratory trials. Config 1 (400/0) follows closely, while Config 3 (200/200) and Config 4 (0/400) exhibit significantly lower performance, suggesting that too many exploratory trials degrade the model’s effectiveness.



\subsection{Details of Used VLA Architecture}
\label{sec:app-vla-architecture}

We utilized $\pi_0$ as the backbone of the VLA Architecture. Here, we present more details about $\pi_0$ design.

$\pi_0$ builds on a pre-trained vision–language model (VLM) backbone (specifically, PaliGemma), inheriting its rich semantic and visual reasoning capabilities. Image observations from the robot’s cameras are encoded into embeddings that share the same space as language tokens. 
To turn the VLM into a controller, $\pi_0$ augments this backbone with two robotics-specific components: (1) proprioceptive state inputs (e.g., joint angles) and (2) continuous action outputs. Both are projected into the transformer’s token sequence. Crucially, $\pi_0$ uses a separate “action expert” — a distinct set of transformer weights — to process these robotics tokens. 
Rather than discretizing actions, $\pi_0$ employs conditional flow matching to model the distribution of continuous action chunks. At each timestep, the model predicts a short sequence (chunk) of future actions by learning a denoising vector field that is integrated via a fixed-step Euler solver at inference. This diffusion-style approach gives $\pi_0$ high precision and the ability to represent complex, multimodal action distributions needed for dexterous tasks.

$\pi_0$ is trained on data from multiple robot platforms (single-arm, dual-arm, mobile bases), each with different action/state dimensions. To accommodate this diversity, all configuration and action vectors are zero-padded to a common size, and missing camera slots are masked. This cross-embodiment strategy lets a single $\pi_0$ model generalize across varied hardware setups without per-robot specialization.

Regarding \textit{RoboChemist}, we introduce an image containing a visual prompt, which consists of annotated points or bounding boxes indicating keypoints and objects. These visual prompts are generated based on the objectives of primitive chemical experiments and corresponding safety guidance. Under the guidance of these visual prompts, the robot is able to perform experimental tasks with greater accuracy and enhanced safety. Specifically, we directly annotate the visual prompts on the RGB image and include them as an additional input channel to the image encoder in the VLA model. We also explicitly indicate in the instruction whether a given image includes a visual prompt. Furthermore, to ensure instruction diversity, we employ GPT-4o to generate multiple variations of natural language instructions for each primitive task, as detailed below.

\begin{enumerate}
    \item \textbf{Grasping a Glass Rod:}
    \begin{tcolorbox}[enhanced, breakable, fontupper=\ttfamily]
1) "In the image input, the last image is used as a reference image, with the [COLOR] target point being the location where the robotic gripper grasps the glass rod. The [COLOR] bounding box surrounds the glass rod, indicating the region of interest. Using the right arm of the robotic arm, carefully grasp the glass rod and lift it gently."

2) "In the last image of the input sequence, the [COLOR] target point indicates the designated grasp location on the glass rod. The [COLOR] bounding box encloses the glass rod, marking the region to focus on. Using the right manipulator, precisely approach and grasp the rod at the indicated point, ensuring a secure and stable hold."

3) "Refer to the last image provided, in which the [COLOR] target point specifies the grasp location on the glass rod. The [COLOR] bounding box highlights the glass rod’s region of interest. The right robotic arm should be used to perform a precise and stable grasp at the indicated position."

4) "As shown in the final image input, the [COLOR] point represents the target location for grasping the glass rod. The [COLOR] bounding box defines the region of the glass rod. Utilize the right arm of the robot to perform a careful and firm grasp at this location, ensuring the rod is securely held."

5) "The last image in the input sequence provides the reference for grasping, with the [COLOR] point indicating the target position on the glass rod. The [COLOR] bounding box clearly identifies the region of the glass rod. The task is to control the robot’s right arm to grasp the rod at this point and maintain a stable grip."
    
    \end{tcolorbox}
    
    \item \textbf{Heating Platinum Wire:}
    \begin{tcolorbox}[enhanced, breakable, fontupper=\ttfamily]
    1) "In the image input, the last image is used as a reference image, with the [COLOR] target point for the platinum wire head to extend into the alcohol burner flame. Using the right robotic arm, hold the platinum wire and carefully extend it into the outer flame of the Bunsen burner until it glows red-hot."

    2) "The final image in the input serves as a reference, with the [COLOR] marker specifying the target location for introducing the platinum wire tip into the Bunsen burner flame. The right robotic manipulator is used to securely hold the wire and extend it into the outer flame region until red-hot.”

    3) "Refer to the last input image, where the [COLOR] target point marks the location for inserting the platinum wire tip into the flame of the Bunsen burner. The robot's right arm should be used to hold the wire and steadily guide it into the outer flame until visible incandescence is achieved.”

    4) "In the last image provided, the [COLOR] point indicates the desired position for extending the platinum wire tip into the Bunsen burner flame. The right robotic arm is tasked with holding the wire and positioning it within the outer flame zone until it becomes red-hot.”

    5) "The [COLOR] marker in the final input image denotes the target region for positioning the platinum wire head within the Bunsen burner flame. The right arm of the robot is employed to grasp and insert the wire into the outer flame carefully, heating it until it glows red.”
    
    \end{tcolorbox}
    
    \item \textbf{Inserting Platinum Wire into Solution:}
    \begin{tcolorbox}[enhanced, breakable, fontupper=\ttfamily]
    
1) "In the last image, the [COLOR] bounding box surrounds the beaker and the [COLOR] target point marks the liquid level inside it. Using the right robotic arm, carefully grasp the platinum wire and gently extend it into the beaker to dip it into the liquid.”

2) "The last image in the input sequence serves as a reference, where the [COLOR] bounding box outlines the beaker and the [COLOR] marker denotes the target liquid level inside it. The right robotic arm is used to securely hold the platinum wire and gently insert it into the liquid up to the specified depth.”

3) "As shown in the final input image, the [COLOR] bounding box highlights the beaker and the [COLOR] target point indicates the liquid surface level. The robot’s right manipulator is employed to grasp the platinum wire and immerse it into the liquid to the designated level.”

4) "Refer to the last image in the input, where the [COLOR] bounding box encloses the beaker and the [COLOR] target point represents the desired immersion depth corresponding to the liquid level. The platinum wire is held by the right robotic arm and is carefully dipped into the liquid accordingly.”

5) "In the final image of the input, the [COLOR] bounding box frames the beaker and the [COLOR] point indicates the liquid level to which the platinum wire should be submerged. The right robotic arm is used to delicately lower the wire into the beaker until the required depth is reached.”
    \end{tcolorbox}
    
    \item \textbf{Pouring Liquid:}
    \begin{tcolorbox}[enhanced, breakable, fontupper=\ttfamily]
1) "In the last image, the [COLOR] bounding box around the left beaker and the [COLOR] bounding box around the right beaker are shown, each containing its respective [COLOR] grasp point (one on the left beaker, one on the right). The [COLOR] point on the right beaker indicates the position for the robotic arm to grasp the beaker. With the left arm holding the left beaker (at the [COLOR] left‐beaker point inside its [COLOR] bounding box) and the right arm grasping the right beaker (at the [COLOR] right‐beaker point inside its [COLOR] bounding box), carefully pour the liquid from the right beaker into the left until fully transferred.”

2) "The last image serves as a reference, showing the [COLOR] bounding box around the left beaker, the [COLOR] bounding box around the right beaker, and their corresponding [COLOR] grasp points. The [COLOR] marker on the right beaker denotes the designated grasp location. The robotic system uses its left arm to stabilize the left beaker (holding it at the [COLOR] left‐beaker point) while the right arm lifts and tilts the right beaker (grasped at the [COLOR] right‐beaker point) to transfer the liquid completely into the left one.”

3) "In the final image, you can see the [COLOR] bounding box around the left beaker and the [COLOR] bounding box around the right beaker, each highlighting a [COLOR] grasp point. The [COLOR] point on the right beaker indicates where to grasp. The robot uses its left manipulator to hold the left beaker steady (at its [COLOR] point) while the right manipulator grasps the right beaker (at the [COLOR] point) and pours its contents into the left until the transfer is complete.”

4) "Refer to the last image, which shows a [COLOR] bounding box around the left beaker and a [COLOR] bounding box around the right beaker, each with an associated [COLOR] point. The [COLOR] marker on the right beaker identifies the designated grasp position. The dual‐arm system coordinates both manipulators—left for stabilizing the receiving beaker (at the [COLOR] left‐beaker point) and right for pouring (at the [COLOR] right‐beaker point)—to execute a complete liquid transfer.”

5) "In the final image of the input, the [COLOR] bounding box around each beaker and their corresponding [COLOR] grasp points are displayed (one on the left, one on the right). The [COLOR] point on the right beaker indicates where to grasp. The robot is instructed to use its left arm to hold the left beaker steady (at the [COLOR] left‐beaker point) and its right arm to grasp the right beaker (at the [COLOR] right‐beaker point), then carefully pour the liquid into the left beaker until the transfer is complete.”

    \end{tcolorbox}
    
    \item \textbf{Stirring Solution with a Glass Rod:}
    \begin{tcolorbox}[enhanced, breakable, fontupper=\ttfamily]
1) "In the last image, the [COLOR] bounding box highlights the beaker. Use the right arm to grasp the spatula and stir inside that box."

2) "The final image shows a [COLOR] box around the beaker. Command the right arm to pick up the spatula and stir within this box."

3) "In the last frame, a [COLOR] bounding box encloses the beaker. Have the right manipulator grasp the spatula and stir inside that region."

4) "Referencing the last image, you’ll see a [COLOR] box around the beaker. Instruct the right arm to hold the spatula and stir within the boxed area."

5) "In the final image, a single [COLOR] bounding box marks the beaker. Use the right arm to grasp the spatula and stir inside the box."

    \end{tcolorbox}
    
    \item \textbf{Transferring Solid:}
    \begin{tcolorbox}[enhanced, breakable, fontupper=\ttfamily]
1) "In the last image, the [COLOR] boxes highlight the left (solid) and right (liquid) cups, each with a [COLOR] point. The left [COLOR] point is where to scoop solid; the right [COLOR] point marks the liquid surface. Use the right arm to grasp the spatula, scoop at the left cup’s [COLOR] point, and pour into the right cup’s [COLOR] point."

2) "The last image shows [COLOR] boxes around both cups and [COLOR] markers for scoop and pour points—the left for solid, the right for liquid. With the right arm, grasp the spatula, scoop at the left [COLOR] point, then deposit into the right [COLOR] point."

3) "In the final image, two [COLOR] boxes enclose the cups, each with a [COLOR] point: left for scooping solid, right for the liquid level. The right manipulator holds the spatula, scoops at the left [COLOR] point, and pours at the right [COLOR] point."

4) "Refer to the last image’s [COLOR] boxes and [COLOR] points—left at the solid’s scoop location, right at the liquid level. The right arm grabs the spatula, scoops at the left [COLOR] point, and delivers into the right [COLOR] point."

5) "In the final image, [COLOR] boxes and points mark the scoop (left) and pour (right) locations. Use the right arm to pick up the spatula, scoop at the left [COLOR] point, and transfer into the right [COLOR] point."

    \end{tcolorbox}
    
    \item \textbf{Pressing a Button:}
    \begin{tcolorbox}[enhanced, breakable, fontupper=\ttfamily]
    1) "In the image input, the last image is used as a reference image, with the [COLOR] target point indicating the location of the switch. Using the right arm of the robotic arm, carefully extend to the red switch and flick it to the left to turn it on."

    2) "In the final image of the input, the [COLOR] target point indicates the location of the switch. Using the right robotic arm, the system carefully extends toward the switch and flicks it to the left to activate it.”

    3) "The last image in the input sequence serves as a reference, where the [COLOR] marker denotes the switch position. The right manipulator is employed to approach the switch and toggle it leftward to turn it on.”

    4) "Refer to the last image in the input, where the [COLOR] point marks the switch location. The robot’s right arm is tasked with extending to the switch and flipping it to the left to power it on.”

    5) "In the final reference image, the [COLOR] marker identifies the location of the switch. The robotic system extends its right arm to engage the switch by flicking it to the left, thereby switching it on.”
    
    \end{tcolorbox}
\end{enumerate}

\subsection{Error Breakdown}
\label{sec:app-error-breakdown}

We identify core failure sources across five modules: task decomposition, visual prompting, outcome monitoring, VLA violating prompt constraints, and VLA action failure with no recovery. Figure~\ref{fig:app-breakdown} summarizes statistics from 20 failure cases. The most common errors arise from visual prompt misalignment in cluttered scenes, followed by VLM monitoring and VLA constraint violations. We are addressing these issues by incorporating \textbf{ multi-view prompt validation},\textbf{ sensor-assisted monitoring}, and \textbf{stricter prompt supervision during training}, leading to gains in robustness and task success.

\subsection{Generalization on environment, and embodiment}
\label{sec:app-gen-env-embodiment}
\begin{figure}[h]
  \centering
   \includegraphics[width=1.0\linewidth]{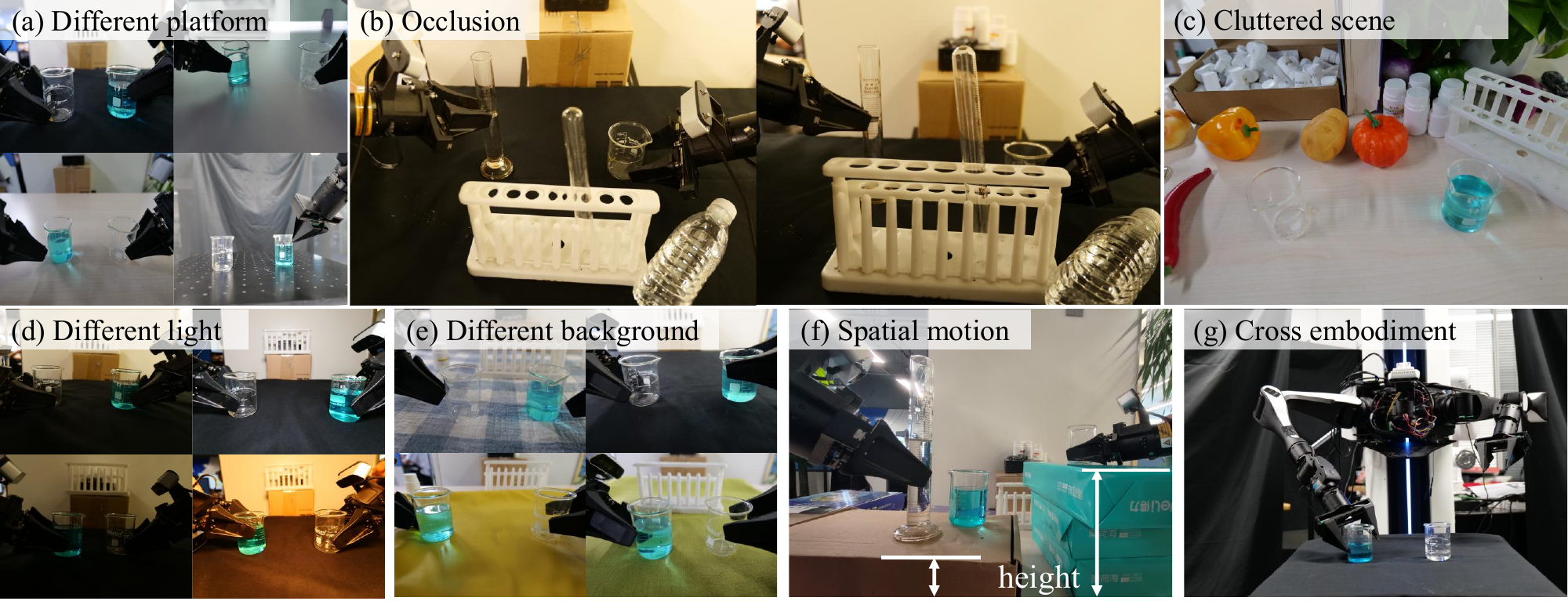}
  \caption{Visualization of primitive task generalization.}

   \label{fig: 123}

\end{figure}

\begin{figure}[t]
  \centering
  \begin{minipage}[t]{0.49\linewidth}
    \centering
    \includegraphics[width=\linewidth]{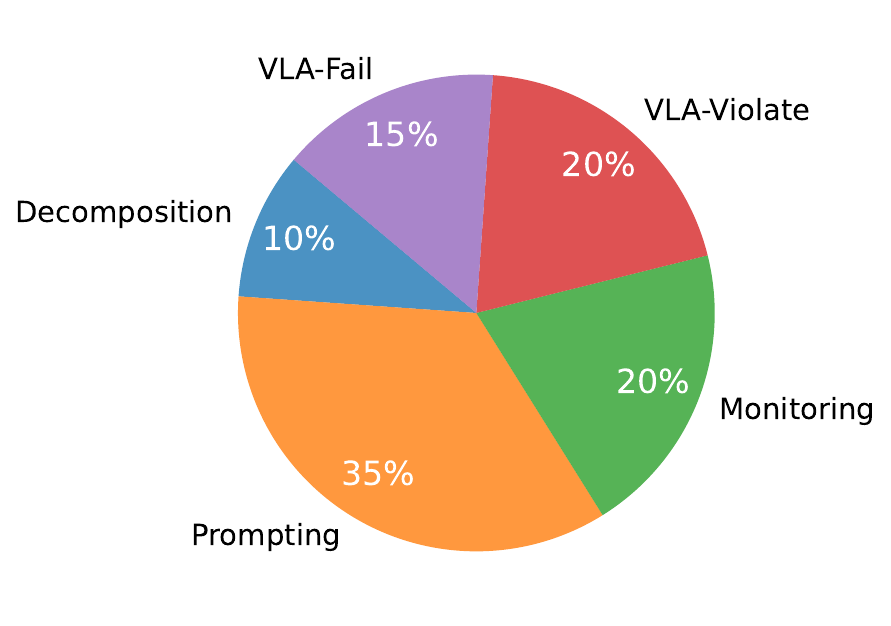}
    \caption{Error breakdown.}
    \label{fig:app-breakdown}
  \end{minipage}\hfill
  \begin{minipage}[t]{0.49\linewidth}
    \centering
    \includegraphics[width=\linewidth]{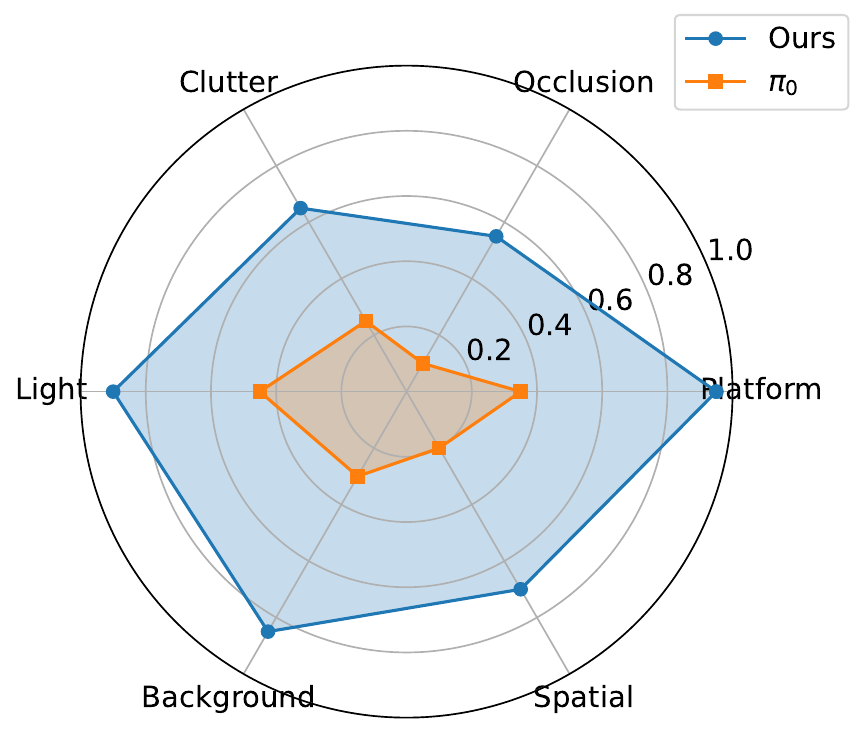}
    \caption{Generalization evaluation.}
    \label{fig:app-lidar}
  \end{minipage}
\end{figure}

We evaluated \textit{RoboChemist}’s generalization across 7 challenging variations, including different platforms, occlusions, cluttered scenes, lighting conditions, table textures, spatial height changes, and embodiments (Figure~\ref{fig: 123}). We evaluated the \textit{Pour Liquid} task and observed strong performance enabled by VLM-based closed-loop feedback and visual prompting (Figure~\ref{fig:app-lidar}).

\end{appendices}


\end{document}